\documentclass[preprint,12pt,authoryear]{elsarticle}
\usepackage{amssymb}
\usepackage{amsmath}
\usepackage{interval}
\intervalconfig{soft open fences, separator symbol=;,}
\usepackage{algorithm}
\usepackage{algorithmic}
\usepackage{graphicx}
\usepackage{url}
\usepackage{booktabs}
\usepackage{longtable}
\usepackage{siunitx}
\usepackage{diffcoeff}

\usepackage{lineno}

\newcommand{\ui}[2]{#1 _{\text{#2}}}
\newcommand{\uis}[3]{#1 _{\text{#2}, #3}}

\renewcommand{\algorithmicrequire}{\textbf{Input:}}
\renewcommand{\algorithmicensure}{\textbf{Output:}}

\begin{document}

\begin{frontmatter}


	\title{Change-Point Detection in Industrial Data Streams based on Online Dynamic Mode Decomposition with Control}

	\author[aff1,aff2]{Marek Wadinger\corref{cor1}}
	\ead{marek.wadinger@stuba.sk}
	\cortext[cor1]{\textit{URL:} \url{uiam.sk/~wadinger} (Marek Wadinger)}

	\author[aff1]{Michal Kvasnica}
	\ead{michal.kvasnica@stuba.sk}

	\author[aff2]{Yoshinobu Kawahara}
	\ead{kawahara@ist.osaka-u.ac.jp}

	\affiliation[aff1]{organization={Institute of Information Engineering, Automation and Mathematics, Slovak University of Technology in Bratislava},
		addressline={Radlinskeho 9},
		postcode={812 37},
		state={Bratislava},
		country={Slovakia}
	}

	\affiliation[aff2]{organization={Machine Learning \& Systems Laboratory, Osaka University},
		addressline={1--5 Yamadaoka},
		postcode={565--0871},
		state={Suita, Osaka},
		country={Japan}
	}

	\begin{abstract}
		We propose a novel change-point detection method based on online Dynamic Mode Decomposition with control (ODMDwC). Leveraging ODMDwC's ability to find and track linear approximation of a non-linear system while incorporating control effects, the proposed method dynamically adapts to its changing behavior due to aging and seasonality. This approach enables the detection of changes in spatial, temporal, and spectral patterns, providing a robust solution that preserves correspondence between the score and the extent of change in the system dynamics. We formulate a truncated version of ODMDwC and utilize higher-order time-delay embeddings to mitigate noise and extract broad-band features. Our method addresses the challenges faced in industrial settings where safety-critical systems generate non-uniform data streams while requiring timely and accurate change-point detection to protect profit and life. Our results demonstrate that this method yields intuitive and improved detection results compared to the Singular-Value-Decomposition-based method. We validate our approach using synthetic and real-world data, showing its competitiveness to other approaches on complex systems' benchmark datasets. Provided guidelines for hyperparameters selection enhance our method's practical applicability.

	\end{abstract}


	\begin{keyword}
		Change-point detection
		\sep{Dynamic mode decomposition}
		\sep{Control systems}
		\sep{Online learning}
		\sep{Data streams}
	\end{keyword}

\end{frontmatter}


\section{Introduction}\label{sec:introduction}
\subsection{Context and Motivation}
Many industrial systems are safety-critical, where process monitoring is essential to protect both profit and life. These systems operate at various operating points, often driven by control to meet desired production goals. However, environmental fluctuations and varying component quality may lead to unexpected and persistent changes in system behavior. These events can jeopardize optimal operations, accelerate wear and tear, and occasionally result in catastrophic consequences, such as equipment damage, production loss, or even human casualties.

Monitoring abrupt and gradual changes in system behavior is crucial for ensuring system reliability and safety, a task known as change-point detection (CPD). Traditional system monitoring methods, like Statistical Process Control (SPC), rely on the assumption that data are independent and identically distributed (i.i.d.), which is often not the case in industrial systems. Industrial process data are typically correlated and non-stationary, complicating the application of SPC methods. While SCADA-based systems do not rely on i.i.d. and use static thresholds to detect changes, they cannot adapt to the dynamic changes in system behavior due to aging or environmental shifts.

Conventionally, offline machine learning (ML) methods are employed to identify macro-scale events in complex, high-dimensional dynamical systems. These methods depend on extensive historical data and require offline training to detect system behavior changes. Although supervised ML methods with annotated data offer high accuracy, they often fail in new contexts or when encountering unexpected data patterns. Moreover, these methods are impractical for existing industrial infrastructures where storing data on a large scale is infeasible due to undeveloped database infrastructure, and direct integration with ongoing data exchange services is necessary.

Indeed, industrial data are streamed and arrive at non-uniform rates, challenging methods dependent on uniform sampling. For instance, \citet{Liu2023} simultaneously detecting change points and anomalies by leveraging the rate of change, and \citet{Fathy2019} using cooperative adaptive filtering for change detection in wireless sensor networks both assume i.i.d. Therefore, these methods may not be suitable for non-uniform data streams typical in industrial settings.

Additionally, sequential data in industrial systems comprise distinct components: linear, seasonal, cyclic, regressions, interventions, and errors. Effective CPD methods must adapt to these changing conditions over the system's lifetime. Further consideration must be given to the type of change point we wish to detect. Variance change points affect more significant segments of time series, while additive change points are sudden pulses that die out quickly, and innovational change points are followed by gradual decay back to the original time series \citep{Srivastava2017}.

This paper provides a unified methodology to answer questions that arise:
\begin{itemize}
	\item Can we detect changes in system behavior using streaming data?
	\item How can changes be detected in the presence of non-stationarity?
	\item Can we adapt to changing system behavior to maintain validity over the system's lifetime?
\end{itemize}

\subsection{Related Work}
Online CPD methods address these questions, which are central for real-time monitoring in safety-critical industrial systems. Unlike offline CPD methods, which typically provide robust detection with significant delay, online CPD offers real-time solutions essential for timely intervention and maintenance planning. Indeed, offline methods often retrospect the historical data and detect changes in the system behavior after the event has occurred. In some settings, this may be acceptable, and favorable properties of offline methods might be enjoyed. For example, \citet{Liu2022} developed a CPD framework using a dynamic Bayesian network model to capture causal relationships between variables, enhancing interpretability and credibility. However, in industrial environments, real-time monitoring is imperative to prevent production losses or catastrophic outcomes such as equipment failure.

Self-supervised approaches are frequently used in online CPD due to the impracticality of obtaining real-time ground truth annotations. However, they still rely on annotations to create a feedback loop. These may be available from other online subsystems that may provide labels for continually supervised approaches to CPD, such as~\cite{Korycki2021}. In most cases, the supervisory information is exploited directly from the raw unlabeled data, showing improved generalization abilities~\citep{Zhang2024}.  

\citet{Chu2022} propose a sequential nearest neighbor search for high-dimensional and non-Euclidean data streams. A stopping rule is proposed to alert detected CP as soon as it occurs while providing a maximum boundary on a number of false positives. Despite its innovation, the sensitivity of nearest neighbors methods to varying data densities and computational expense in high-dimensional spaces restricts its real-time applicability.

\citet{Gupta2022} proposed a three-phase architecture for real-time CPD using autoencoders (AE). However, the necessary preprocessing steps, such as shifting and scaling, require assumptions about data distribution that are often unknown in streaming scenarios. Additionally, recursive singular spectrum analysis, employed in the architecture, may impose significant computational overhead in the case of high-dimensional data.

\citet{Bao2024} proposes feature decomposition and contrastive learning (CoCPD) for industrial time series to detect both abrupt CPs and subtle changepoints. By isolating predictable components from residual terms, this method improves detection accuracy in detecting subtle changes. Contrastive learning methods rely on constructing negative samples to increase the energy of the change points and decrease the energy of the stationary operation data. Nevertheless, this is one of the main bottlenecks of contrastive learning methods. Since changes are unpredictable events that differ in sources and nature, it is challenging to generate negative samples to capture this variability as the prior information on the magnitudes and timing distributions are unknown, and the space of negative samples is therefore unbounded.

Established statistical CPD methods promote interpretability while remaining highly competitive.~\citet{Rajaganapathy2022} introduced a Bayesian network-based CPD method, which is able to capture CPs characterized by step change leveraging causal relationships between the variables. Nevertheless, as we will explore soon, the change point may be characterized by a change in dynamics, which is better captured in the frequency domain rather than in the time domain.


Another common practice in CPD is to compare past and future time series intervals using a dissimilarity measure, triggering alarms when intervals are sufficiently different. Statistical CPD methods usually compare the relative statistical differences between time intervals to identify change points (CPs). Temporal properties such as data distribution and time series models should be accurately modeled in advance to obtain more precise statistical metrics for evaluating interval homogeneity. Methods in this group define this dissimilarity measure based on the difference in distribution of the two intervals. For instance, CUSUM and related methods \citet{Ye2023} track changes in the parameter of a chosen distribution, and the generalized likelihood ratio (GLR) procedure \citet{Xie2013, Korycki2021} monitors the likelihood that both intervals are generated from the same distribution. Subspace-based methods measure the distance between subspaces spanned by the columns of an observability matrix \citet{Moskvina2003, Kawahara2007} or observe reconstruction error~\citep{DeRyck2021, Bao2024}. Here, we will use subspace-based methods as an umbrella term for decomposition and deep neural network methods, which rely on finding the low-dimensional description of the data.  

\subsection{Subspace-based CPD}\label{sec:subspace-based-cpd}
The main advantages of subspace-based approaches include the absence of distributional assumptions and their ability to extract complex dynamical features from data efficiently. For example, \citet{Hirabaru2016} might be a powerful example of cost-effectiveness in high-dimensional systems. The authors find a 1D subspace within multidimensional data and apply efficient univariate CPD, expanding their applicability to multivariate scenarios. This operates under the same assumption as \citet{Fathy2019} that the measurements are closely related and enable 1D representation to capture the signal from the noise. Nevertheless, these approaches are not suitable for complex systems characterized by multiple weakly related quantities whose behavior cannot be captured solely by the largest eigenvalue.

While some ML-based transformers demonstrate the ability to adapt and generalize to new data while retaining useful information over prolonged deployments \citep{Corizzo2022}, they lack guarantees that the CPD score accurately reflects the actual dissimilarity between intervals. This issue, highlighted by \citet{DeRyck2021}, can lead to misjudgments about the severity of change points and result in poor decision-making. Additionally, subspace-based methods are sensitive to hyperparameter choices, often lacking informed guidance.

Subspace-based methods address these limitations by monitoring whether incoming data aligns with the null space of the reference state's observability matrix, effectively identifying new operating states \citep{Dohler2013, Ye2023}. \citet{Xie2013} leverage this principle in the MOUSSE (Multiscale Union of Subsets Model) algorithm, which tracks dynamic submanifolds in high-dimensional noisy data using a sequential generalized likelihood ratio procedure for CPD.

Numerous theoretical studies support the optimality of subspace-based methods in CPD. For instance, \citet{Ye2023} derive an exact subspace-CUSUM procedure and characterize average run length (ARL) and Type-II error probability using asymptotic random matrix theory, optimizing metrics such as expected detection delay (EDD). Similarly, \citet{Garreau2018} demonstrate that a kernel change-point algorithm can, with high probability, correctly identify the number and location of change points under well-chosen penalties and estimate the change-point location at the optimal rate.

Successful practical applications complement the solid theoretical foundation of these methods. For instance, \citet{Hosur2019} address the need for near real-time detection of changes in power systems' working conditions with a sequential detection algorithm based on stochastic subspace state-space identification, utilizing output-only covariance-based subspace identification with Hankel matrices. \citet{Dohler2013} propose a robust residual function for detecting changes in the eigenstructure of linear time-invariant systems for vibration monitoring. \citet{He2019} introduce ADMOST, an online subspace tracking framework similar to online SVD updated without increasing the rank, displaying its applicability to real-time UAV flight data, where anomaly detection and mitigation are required.

\subsection{DMD-based CPD}
In both autonomous and controlled dynamical systems, change points may be characterized by shifts in dynamics that are more effectively captured in the frequency domain rather than the time domain \citep{DeRyck2021, Gupta2022}. Addressing this issue requires decomposing a time series into its dominant frequency components, which are described by oscillations and magnitudes. For example, \citet{DeRyck2021} combined detection in both domains using two autoencoders in the TIRE method, leveraging discrete Fourier Transformation to extract spectral information. Similarly, \citet{Gupta2022} utilized recursive singular spectrum analysis in preprocessing within an autoencoder-based CPD framework to decompose time series into dominant frequency components. However, this approach requires retraining the model after each predicted change point, which is computationally expensive and unsuitable for real-time applications.

Dynamic Mode Decomposition (DMD) emerges as a suitable method for CPD in both the time and frequency domain. DMD is a data-driven technique that decomposes a time series into its dominant frequency components (modes), described by their oscillation and magnitudes. By approximating the dynamical system through a linear combination of these modes, DMD facilitates interpretable CPD concerning system dynamics. It allows for monitoring changes in spatial features and system dynamics and detecting changes arising from environmental factors.

Supporting this claim, \citet{Prasadan2020} applied DMD to a data matrix composed of linearly independent, additive mixtures of latent time series, focusing on missing data recovery. They demonstrated that hankelized DMD, a higher-lag extension of DMD, could unmix signals, revealing them better in noise and offering superior reconstruction compared to Principal Component Analysis (PCA) and Independent Components Analysis (ICA).

In another study, \citet{Srivastava2017} introduced an innovative offline algorithm leveraging DMD to detect variance change points iteratively. Their method integrates a data-driven dynamical system with a local adaptive window guided by a variance descriptor function, facilitating the identification of change points at various scales. By employing sequential hypothesis testing and a dynamic window mechanism, the method dynamically adjusts the window's location and size to detect changes in variance. This offline algorithm performs multiple passes over the data first to identify the longest stationary segments and then detect variance change points, making it unsuitable for real-time applications.

Similarly, \citet{Gottwald2020} focused on detecting transient dynamics and regime changes in time series using DMD. They argued that transitions between different dynamical regimes are often reflected in higher-dimensional space, followed by relaxation to lower-dimensional space. They proposed using the reconstruction error of DMD to monitor a time series' inability to resolve fast relaxation towards the attractor and the system dynamics' effective dimension.

\subsection{Research Objective and Contributions}
This paper proposes a novel online change-point detection (CPD) method based on truncated online Dynamic Mode Decomposition (DMD) with control. We leverage DMD's capability to decompose time series into dominant frequency components and incorporate control effects to adapt to changing system behaviors. The proposed method detects abrupt changes in system behavior, considering both the time and frequency domains. We demonstrate the effectiveness of this approach on real-world data streams, showing that it is highly competitive or superior to other general CPD methods in terms of detection accuracy on benchmark datasets.

The significance of this work is underscored in industrial settings where complex dynamical systems are challenging to describe, data arrive at non-uniform rates, and real-time assessment of changes is crucial to protecting both profit and life.

The main contributions of this paper are:
\begin{itemize}
	\item Formulation of a truncated version of online DMD with control for tracking system dynamics.
	\item Utilization of higher-order time-delay embeddings in streamed data to extract broad-band features.
	\item Demonstration that using the DMD improves the detection accuracy compared to SVD-based CPD methods.
	\item Analysis of the correspondence between increases in detection statistics and the actual dissimilarity of compared intervals.
	\item Validation of the proposed method's effectiveness on real-world data from a controlled dynamical system.
	\item Provision of intuitive guidelines for selecting hyperparameters for the proposed method.
\end{itemize}

\section{Preliminaries}\label{sec:preliminaries}
This section presents the theoretical background of the proposed method. We start with the definition of Dynamic Mode Decomposition (DMD) and its online and extended online versions. We then describe how to utilize the online Singular Value Decomposition (SVD) algorithm, which enables finding lower rank representation less expensively. Finally, we present the proposed method for truncating the DMD matrix to a lower rank online.

\subsection{DMD}\label{sec:dmd}
Dynamic Mode Decomposition (DMD), introduced in \citet{Schmid2010}, is a technique with broad application in data sequence analysis. The use cases span discriminating dominant signal and noise components from high-dimensional measurements, revealing coherent structures, and modeling dynamic behavior via system identification. The DMD was found to be closely related to Koopman theory by \citet{Rowley2009}, revealing perhaps the most interesting property of representing a non-linear system as a set of linear governing equations, which enabled its combination with nominal MPC and other techniques where optimization problem could be significantly simplified by finding linear representation of the system albeit increased dimensionality of the model. Various modifications of DMD further broadened its utilization and underpinned its essential place in system identification and control theory \citep{Schmid2022}.

The DMD algorithm aims to find the optimal linear operator \(A\) that advances the snapshot matrix in time; mathematically, the optimal linear operator \(A\) is defined as
\begin{equation}\label{eq:best-fit-operator}
    A = \mathrm{argmin} ||X^\prime - AX||_F = X^\prime X^+,
\end{equation}
where matrices \(X \in \mathbb{C}^{m \times n}\) and \(X^\prime \in \mathbb{C}^{m \times n}\) represent \(n\) consecutive snapshot pairs \({\{x(t_i), x(t_i^\prime )\}}^n_{i=1}\), where \(t_i^\prime = t_i + \Delta t_i\) and \(X^+\) is Moore-Penrose pseudinverse of \(X\).

\citet{Tu2013} proposed an exact algorithm for solving~\eqref{eq:best-fit-operator}, that does not rely on the assumption of uniform sampling, enabling its usage in industrial data streams. While enabling irregular sampling, time steps \(\Delta t_i\) must be sufficiently small to capture the highest frequency dynamics.

\subsection{Algorithm for DMD}
DMD utilizes the computationally efficient singular value decomposition (SVD) of \(X\) to provide a low-rank representation of high-dimensional systems.

\begin{equation}\label{eq:svd}
    X = \tilde{U} \tilde{\Sigma} \tilde{V}^\top,
\end{equation}
where \(\tilde{U} \in \mathbb{C}^{m \times r}\) are proper orthogonal decomposition (POD) modes, \(\tilde{\Sigma} \in \mathbb{C}^{r \times r}\) are the singular values, and \( \tilde{V} \in \mathbb{C}^{n \times r}\) are right orthogonal singular vectors. Rank \(r \leq m\) denotes either the full or the approximate rank of the data matrix \(X\).

Employing~\eqref{eq:svd} we may rewrite~\eqref{eq:best-fit-operator} as
\begin{equation*}
    A = X^\prime \tilde{V} \tilde{\Sigma}^{-1} \tilde{U}^\top,
\end{equation*}

and project \(A\) onto the POD modes \(\tilde{U}\) to obtain low rank representation \(\tilde{A}\) as
\begin{equation}\label{eq:projected-operator}
    \tilde{A} = \tilde{U}^\top A \tilde{U} = \tilde{U}^\top X^\prime \tilde{V} \tilde{\Sigma}^{-1}.
\end{equation}

Unlike SVD, which focuses on spatial correlation and energy content, DMD incorporates temporal information via spectral decomposition of the matrix \(\tilde{A}\) as
\begin{equation}\label{eq:dmd-modes}
    \tilde{A} W = W \Lambda.
\end{equation}

where diagonal elements of \(\Lambda \), \({\{\lambda}_0, {\lambda}_1, \ldots, {\lambda}_r\} \), are the DMD eigenvalues, and columns of \(W\) are the DMD modes.

Projection onto POD modes in~\eqref{eq:projected-operator} preserves the non-zero eigenvalues of the full matrix \(A\), removing the necessity of working with the high-dimensional \(A\) matrix in~\eqref{eq:dmd-modes}.

DMD modes represent linear combinations of POD mode amplitudes with consistent linear behavior over time, offering insights into temporal evolution, thus combining the advantages of SVD for spatial dimensionality reduction and FFT for identifying temporal frequencies. Each DMD mode is linked to a specific eigenvalue \(\lambda = a + ib\), indicating growth or decay rate \(a\) and oscillation frequency \(b\).

Therefore, DMD not only reduces dimensionality but also models the evolution of the modes in time, enabling its usage for prediction\citep{Brunton2022}. Indeed, the operator \(A\) then represents a linear time-invariant system

\begin{equation*}
    X^\prime = AX.
\end{equation*}

Lastly, to reconstruct the full-dimensional DMD modes \(\Phi \) from reduced DMD modes \(W\) we use time-shifted snapshot matrix \(X^\prime \) obtaining
\begin{equation}\label{eq:full-dmd-modes}
    \Phi = X^\prime \tilde{V} \tilde{\Sigma}^{-1} W.
\end{equation}

\citet{Tu2013} has shown correspondence between DMD modes and eigenvectors of the full matrix \(A\) as
\begin{align*}
    A\Phi
     & = (X^\prime \tilde{V} \tilde{\Sigma}^{-1} \tilde{U}*) (X^\prime \tilde{V} \tilde{\Sigma}^{-1} W)
    \tilde{A}                                                                                           \\
     & = X^\prime \tilde{V} \tilde{\Sigma}^{-1} \tilde{A} W                                             \\
     & = X^\prime \tilde{V} \tilde{\Sigma}^{-1} W \Lambda                                               \\
     & = \Phi \Lambda
\end{align*}

In cases where \(n \gg m\), we may seek more efficient reconstruction of the full-dimensional DMD modes using projected modes \(\Phi = \tilde{U} W\) while losing guarantees of finding the exact eigenvectors of \(A\) \citep{Schmid2010}.

\subsection{Online DMD}
In most practical applications, sufficient data may not be available on demand but instead become available in a streaming manner. Moreover, many complex systems in nature or engineered ones exhibit time-varying dynamics, under the influence of environmental or operational factors, that we may wish to track over time to maintain the models' validity. In these relevant cases, we can update the underlying decomposition of the data matrix \(X\) over time.

Recently, an attractive way of updating exact DMD in streaming applications was proposed by \citet{Zhang2019}, providing extensive variations to improve tracking of time-varying dynamics without storing the full data matrix \(X\).

\subsection{Algorithm for online DMD updates}\label{sec:online-dmd-updates}
The initial requirement of online DMD updates in \citet{Zhang2019} is the availability of \(A\). In some instances, we may have recorded (or sufficient time to record) the history of snapshots \(X_k\) up to time step \(k\) enabling initialization of \(A_k\) using the standard DMD algorithm presented in Section~\ref{sec:dmd}. Conversely, initializing \(A_k\) with the identity matrix works well in practice and converges quickly.

In streaming data processing, new pairs of snapshots may become available in real-time or delayed in mini-batches as
\begin{equation}
    \{X_{k : k + c}, X^\prime_{k : k + c}\} = {\{x(t_i), x(t_i^\prime )\}}^c_{i=k}.
\end{equation}

We wish to find an updated matrix \(A_{k+c}\), assuming it is close to \(A_{k}\), enabling the formulation of the problem as recursive least-squares estimation. Under the assumption that the history of snapshots \(X_k\) has rank \(m\), and the matrix \(X_k X_k^\top \) is symmetric and strictly positive definite and has a well-defined inversion, we may rewrite~\eqref{eq:best-fit-operator} as
\begin{equation}\label{eq:pseudo-inverse}
    A_k = X^\prime_k X_k^+ = X^\prime_k X_k^\top {(X_k X_k^\top)}^{-1} = Q_k P_k,
\end{equation}

where \(Q_k\) and \(P_k\) are \(m \times m\) lag covariance matrix and precision matrix respectively, given by
\begin{align}\label{eq:aux-matrices}
    Q_k & = X^\prime_k X_k^\top,   \\
    P_k & = {(X_k X_k^\top)}^{-1}.
\end{align}

The DMD matrix may be updated on new pairs of snapshots \( \{X_{k: k + c}, X^\prime_{k: k + c}\} \) by updating the matrices \(Q_k\) and \(P_k\) as
\begin{align*}
    Q_{k+c}      & = \begin{bmatrix} X^\prime_k  & X^\prime_{k : k + c} \end{bmatrix} \begin{bmatrix} X_{k} & X_{k : k + c} \end{bmatrix}^\top = X^\prime_k X_{k}^\top + X^\prime_{k : k + c} C_{k : k + c} X_{k : k + c}^\top \\
    P^{-1}_{k+c} & = \begin{bmatrix} X_{k} & X_{k : k + c} \end{bmatrix} \begin{bmatrix} X_{k} & X_{k : k + c} \end{bmatrix}^\top = X_{k}X_{k}^\top + X_{k : k + c} C_{k : k + c} X_{k : k + c}^\top,
\end{align*}

where diagonal matrix \(C_{k: k + c}\) holds corresponding weights of samples, desirable in scenarios where multi-fidelity data are available, and external agent defines their fidelity in real-time (e.g.~outlier detector).

The update of \(Q_k\) and \(P_k\) then translates to updated DMD matrix \(A_{k+c}\) as
\begin{equation}
    A_{k+c} = (Q_k + X^\prime_{k : k + c} C_{k : k + c} X_{k : k + c}^\top) {(P_k^{-1} + X_{k : k + c} C_{k : k + c} X_{k : k + c}^\top)}^{-1}
\end{equation}

As both matrices, \(C_{k: k + c}\) \(P_k\) are invertible square matrices due to their properties, the Woodbury formula may be used to compute the inverse of the sum of the matrix and its outer product with a vector obtaining
\begin{equation}\label{eq:precision-matrix-update}
    P_{k+c} = {(P_k^{-1} + X_{k : k + c} C_{k : k + c} X_{k : k + c}^\top)}^{-1} = P_k - P_k X_{k : k + c} \Gamma_{k+c} X_{k : k + c}^\top P_k,
\end{equation}

where

\begin{equation}
    \Gamma_{k : k + c} = {(C_{k : k + c}^{-1} + X_{k : k + c}^\top P_k X_{k : k + c})}^{-1}.
\end{equation}

\(\Gamma_{k : k + c}^{-1}\) is always non-zero due to the positive definiteness of \(P_k\), if for all diagonal elements of \(C_{k : k + c}\) applies \( \forall i : c_{ii} \neq 0 \). The inversion of matrix \(C_{k: k + c}\) can be efficiently computed as residuals of diagonal elements.

The final closed-loop form of the updated DMD matrix is then
\begin{equation}\label{eq:online-dmd-update}
    A_{k+c} = A_k + (X^\prime_{k : k+c} - A_k X_{k+c}) \Gamma_{k+c} X_{k : k+c}^\top P_k,
\end{equation}

where \((X^\prime_{k : k+c} - A_k X_{k+c})\) represents the prediction error. The DMD matrix is updated by adding a term proportional to this error, reflecting the data's covariance structure and variable importance through \(C_{k: k + c}\).

\subsection{Windowed Online DMD}
The DMD updates presented in the previous section enable calibration of the DMD modes in scenarios where snapshots become available over time. Increasing the number of observed snapshots increases the accuracy of identification. However, in time-varying systems, the previous snapshots may become invalid and reduce the validity of the found model. In such cases, it may be desirable to revert the DMD matrix to the state it would have been in if the old snapshots had never been included in the so-called windowed online DMD.

To make DMD matrix forget first snapshots seen \( \{X_{c}, X^\prime_{c}\} = {\{x(t_i), x(t_i^\prime )\}}^c_{i=0}\), we simply use the update formulae from~\eqref{eq:precision-matrix-update} and~\eqref{eq:online-dmd-update} providing negative value of their original weights \(-C_{c}\).

This means that the history of snapshot pairs \({\{x(t_i), x(t_i^\prime )\}}^{k+c}_{i=0}\) must be stored until they are reverted. This window might be significantly smaller than all the previously seen data, saving computational resources and memory.

\subsection{Online DMD with Control}
In industrial automation, complex systems to which external control is applied are of interest. DMD can effectively identify internal system dynamics, subtracting the effect of control input. Perhaps more interestingly, it can also be used to evaluate the effect of control on the system \citep{Proctor2016}. From control theory, the (discrete-time) linear time-varying system can be written as
\begin{equation}\label{eq:linear-system}
    X_{k+1} = A_k X_{k} + B_k \Theta_{k},
\end{equation}
where \(X_k \in \mathbb{R}^{m \times c}\), \(\Theta_k \in \mathbb{R}^{l \times c}\) are the states and control inputs, respectively, \(A_k \in \mathbb{R}^{m \times m}\) is the state matrix, and \(B_k \in \mathbb{R}^{m \times l}\) is the control matrix.

For known control matrix \(B\), the control input may be incorporated into the DMD matrix by simply compensating the output snapshots \(X^\prime_k\) with the control input multiplied by the control matrix \(B\) as
\begin{equation}\label{eq:control-compensation}
    \bar{X}^\prime_k = X^\prime_k - B \Theta_k,
\end{equation}

and use the \(\bar{X}^\prime_k\) in place of \(X^\prime_k\) in the update formulae~\eqref{eq:precision-matrix-update} and~\eqref{eq:online-dmd-update}.

In most scenarios, neither internal structure \(A\) nor the effect of control \(B\) are known. In such cases, the system identification problem may be solved by augmenting the state matrix \(A\) with the control matrix \(B\) as
\begin{equation}\label{eq:augmented-matrix}
    \bar{A}_k = \begin{bmatrix} A_k & B_k \end{bmatrix}, \quad \bar{X}_k = \begin{bmatrix} X_k \\ \Theta_k \end{bmatrix} ,
\end{equation}

where \(\bar{A}_k \in \mathbb{R}^{m \times m + l}\), \(\bar{X}_k \in \mathbb{R}^{m + l \times c}\) are the augumented matrices of \(A_k\) and \(X_k\). We may write~\eqref{eq:linear-system} in the form
\begin{equation*}
    X^\prime_k = \bar{A}_k \bar{X}_k.
\end{equation*}

Similarly to DMD, the matrices \(A_k\) and \(B_k\) may then be found by minimizing the Frobenius norm of \(J_k = \|X^\prime_k - \bar{A}_k \bar{X}_k\|_F\) resulting in the same  formula as in~\eqref{eq:best-fit-operator}
\begin{equation*}
    \bar{A}_k = X^\prime_k \bar{X}_k^+.
\end{equation*}

At time \(k+c\), we incorporate \(c\) new columns into \(\bar{X}_k\) and \(X^\prime_k\), and aim to update \(\bar{A}_{k+c}\) utilizing our prior knowledge of \(\bar{A}_k\). By applying the same method as in Section~\ref{sec:online-dmd-updates}, extending the online DMD to this scenario is straightforward. Specifically, the square matrix \(A_k\) from the DMD is replaced in DMDc with the rectangular matrix \(\bar{A}_k\) defined earlier, and the matrix \(X_k\) in the formulae~\eqref{eq:precision-matrix-update} and~\eqref{eq:online-dmd-update} is substituted with the matrix \(\bar{X}_k\)~\citep{Zhang2019}.

\subsection{Truncating Online DMD with Control}\label{sec:truncating-online-dmd}
Some of the challenges of online DMD proposed in~\citet{Zhang2019} include the lack of robustness to noise, bad scalability, and decreased numerical stability of small eigenvalue updates. To address these issues, we propose modifying the online DMD algorithm that truncates the DMD matrix to a lower rank. Conventionally, this process in batch-trained DMD relies on the truncated singular value decomposition (SVD) method, which is widely used in data analysis to reduce the dimensionality of data while preserving the most essential information. Nevertheless, computing the SVD of the matrix \(X_k\) is computationally expensive and unsuitable for online learning. Instead, we employ online SVD algorithms that perform low-rank updates of the SVD as new snapshots \(X_k\) become available.

We use the algorithm of \citet{Zhang2022}, a modified version of the originally proposed algorithm by \citet{Brand2006}. The main benefit of this modification is the reorthogonalization rule, which prevents erosion of left singular values orthogonality at a reasonable computational cost. For the details on the algorithm, please refer to the original work of author \citep{Zhang2022}. For consistency of nomenclature, we will refer to the SVD decomposition of the augmented matrix \(\bar{X}_k\) as
\begin{equation*}
    \bar{X}_k = \tilde{U}_k \tilde{\Sigma}_k \tilde{V}_k^\top.
\end{equation*}

The new snapshots \(\bar{X}_{k+c}\) may be used for updating the Online SVD as shown in Algorithm~\ref{alg:online-svd-updates}. Old snapshots may be reverted using Algorithm~\ref{alg:online-svd-reverts}.

\begin{algorithm}[H]
    \caption{{Online SVD Updates}}\label{alg:online-svd-updates}
    \begin{algorithmic}[1]
        \REQUIRE{
        \(\tilde{U} \in \mathbb{R}^{m \times r}\),
        \(\tilde{\Sigma} \in \mathbb{R}^{r \times r}\),
        \(\tilde{V} \in \mathbb{R}^{k \times r}\),
        \(\bar{X}_{k+c} \in \mathbb{R}^{m \times c}\),
        \(\ui{q}{u}\),
        \(\ui{\tilde{\bar{X}}}{buff} \in \mathbb{R}^{k \times \ui{q}{u}}\),
        \(\text{tol}\),
        }
        \ENSURE{
        \(\tilde{U}\),
        \(\tilde{\Sigma} \),
        \(\tilde{V}\),
        \(\ui{q}{u}\),
        \(\ui{\tilde{\bar{X}}}{buff} \in \mathbb{R}^{k \times \ui{q}{u}}\),
        }
        \\ \textit{Set}: {
        \(U_0 = \textbf{I}_{r}\);
        \(\tilde{\bar{X}}_k = \tilde{U}^\top \bar{X}_{k+c}\);
        \(E = \bar{X}_{k+c} - \tilde{U} \tilde{\bar{X}}_k\);
        \(E^\prime, \_ \gets \text{qr}(E) \);
        \(\ui{K}{E'E} = E^{\prime \top} E\);
        }
        \IF{\(\ui{K}{E'E} < tol\)}
        \STATE{
            \(\ui{q}{u} = \ui{q}{u} + 1\);
            \(\ui{\tilde{\bar{X}}}{buff}(:, \ui{q}{u}) = \tilde{\bar{X}}_k\);
        }
        \ELSE{}
        \IF{\(\ui{q}{u} > 0\)}
        \STATE{
            \(Y = \begin{bmatrix}
                \tilde{\Sigma}~|~\ui{\tilde{\bar{X}}}{buff}
            \end{bmatrix}\);
        }
        \STATE{
        \(\begin{bmatrix}
            \ui{U}{Y}, \ui{\Sigma}{Y}, \ui{V}{Y}
        \end{bmatrix}\) = \(\text{svd}(Y)\); where \(\ui{U}{Y} \in \mathbb{R}^{r \times r}\),
        \(\ui{V}{Y} \in \mathbb{R}^{r+\ui{q}{u} \times r}\),
        }
        \STATE{
            \(U_0 = U_0\ui{U}{Y}\);
            \(\tilde{\Sigma} = \ui{\Sigma}{Y}\);
            \(V_1 = \ui{V}{Y}(:r, :-1)\);
            \(\tilde{V}_2 = \ui{V}{Y}(r, :-1)\); \\
            \(\tilde{V} = \begin{bmatrix}
                \tilde{V}V_1 \\ V_2
            \end{bmatrix}\);
            \(\tilde{\bar{X}}_k = \ui{U}{Y}^\top \tilde{\bar{X}}_k\)
        }
        \ENDIF{}
        \IF{\(|E^{\prime\top} \tilde{U}(:, 0)| > tol\)}
        \STATE{
            \(E^\prime = E^\prime - \tilde{U}\tilde{U}^\top E^\prime \);
            \(E^\prime, \_ \gets \text{qr}(E^\prime) \);
        }
        \ENDIF{}
        \STATE{
            \(Y = \begin{bmatrix}
                \tilde{\Sigma} & \tilde{\bar{X}}_k \\ 0 & \ui{K}{E'E}
            \end{bmatrix}\);
        }
        \STATE{
            \(\begin{bmatrix}
                \ui{U}{Y}, \ui{\Sigma}{Y}, \ui{V}{Y}
            \end{bmatrix}\) = \(\text{svd}(Y, \text{rank=}r)\); where \(\ui{U}{Y} \in \mathbb{R}^{r \times r}\), \(\ui{V}{Y} \in \mathbb{R}^{r+c \times r}\),
        }
        \STATE{
            \(\tilde{U} = \begin{bmatrix} \tilde{U}~|~E^\prime \end{bmatrix}U_0\ui{U}{Y}\);
            where \(\tilde{U} \in \mathbb{R}^{m \times r}\),
        }
        \STATE{
            \(\tilde{\Sigma} = \ui{\Sigma}{Y}\); where \(\tilde{\Sigma} \in \mathbb{R}^{r \times r}\);
        }
        \STATE{
        \(\tilde{V} = \begin{bmatrix} \tilde{V} & 0 \\ 0 & \textbf{I}_{c + \ui{q}{u} \times c} \end{bmatrix}\ui{V}{Y}\);
        where \(\tilde{V} \in \mathbb{R}^{k+c+\ui{q}{u} \times r}\)
        }
        \STATE{
        \(U_0 = \textbf{I}_{r}\);
        \(\ui{\tilde{\bar{X}}}{buff} = [-]\);
        \(\ui{q}{u} = 0\);
        }
        \ENDIF{}
    \end{algorithmic}
\end{algorithm}

\begin{algorithm}[H]
    \caption{{Online SVD Reverts}}\label{alg:online-svd-reverts}
    \begin{algorithmic}[1]
        \renewcommand{\algorithmicrequire}{\textbf{Input:}}
        \renewcommand{\algorithmicensure}{\textbf{Output:}}
        \REQUIRE{
        \(\tilde{U} \in \mathbb{R}^{m \times r}\),
        \(\tilde{\Sigma} \in \mathbb{R}^{r \times r}\),
        \(\tilde{V} \in \mathbb{R}^{k+c+\ui{q}{r} \times r}\),
        \(\bar{X}_{k+c} \in \mathbb{R}^{m \times c}\),
        \(\ui{q}{r}\),
        \(\text{tol}\),
        }
        \ENSURE{
            \(\tilde{U} \in \mathbb{R}^{m \times r}\),
            \(\tilde{\Sigma} \in \mathbb{R}^{r \times r}\),
            \(\tilde{V} \in \mathbb{R}^{k \times r}\),
            \(\ui{q}{r}\),
        }
        \\ \textit{Set}: {
        \(B = \begin{bmatrix} \mathbf{0}_{k \times c}  \\ \textbf{I}_{c\times c}\end{bmatrix}\);
        \(N = \tilde{V}^\top (:, :c)\);
        \(E = B - \tilde{V} N\);
        \(E^\prime, \_ \gets \text{qr}(E) \);
        \(\ui{K}{E'E} = E^{\prime\top} E\);
        }
        \IF{\(\ui{K}{E'E} < tol\)}
        \STATE{
            \(\ui{q}{r} = \ui{q}{r} + 1\);
        }
        \ELSE{}
        \IF{\(\ui{q}{r} > 0\)}
        \STATE{
            \(B = \begin{bmatrix} \mathbf{0}_{k \times c + \ui{q}{r}}  \\ \textbf{I}_{c+ \ui{q}{r} \times c + \ui{q}{r}}\end{bmatrix}\);
            \(N = \tilde{V}^\top (:, :c + \ui{q}{r})\);
            \(E = B - \tilde{V} N\);
            \(E^\prime, \_ \gets \text{qr}(E) \);
        }
        \ENDIF{}
        \STATE{
            \(\bar{\tilde{\Sigma}} = \begin{bmatrix}
                \tilde{\Sigma} & 0 \\ 0 & \mathbf{0}_{c+\ui{q}{r}}
            \end{bmatrix}\)
        }
        \STATE{
            \(Y = \bar{\tilde{\Sigma}} \left(  \textbf{I}_{r + c + \ui{q}{r}} -
            \begin{bmatrix}
                N \\ \mathbf{0}_{c + \ui{q}{r}}
            \end{bmatrix}
            \begin{bmatrix}
                N \\ \sqrt{1 - N^\top N}
            \end{bmatrix}^\top
            \right)\);
        }
        \STATE{
            \(\begin{bmatrix}
                \ui{U}{Y}, \ui{\Sigma}{Y}, \ui{V}{Y}
            \end{bmatrix}\) = \(\text{svd}(Y)\);
        }
        \STATE{
            \(\tilde{U} = \tilde{U}\ui{U}{Y}(:r, :r)\);
            \(\tilde{\Sigma} = \ui{\Sigma}{Y}(:r, :r)\);
            \(\tilde{V} = (\begin{bmatrix} \tilde{V}~|~E^\prime \end{bmatrix}\ui{V}{Y})(:k, :r)\);
        }
        \STATE{
            \(\ui{q}{r} = 0\);
        }
        \ENDIF{}
    \end{algorithmic}
\end{algorithm}

Further, we propose incorporating the truncation using online SVD into the online DMD algorithm described in Section~\ref{sec:online-dmd-updates}. The truncation of the DMD matrix requires a data transformation step before updating the DMD matrix. This transformation is performed by projecting the snapshots onto the first \(r\) POD modes as
\begin{equation}
    \{\tilde{\bar{X}}_k, \tilde{X}^\prime_k\} = \{\tilde{U}^\top_k\bar{X}_k, \tilde{U}^\top_k(:m, 1:p)X^\prime_k\}.
\end{equation}

We wish to update reduced-order matrix \(\tilde{\bar{A}}_k\), a rectangular matrix of size \(p \times p + q\), where \(r = p + q\), \(p\) is the rank of the reduced-order state matrix \(\tilde{A}_k\) and \(q\) is the rank of the reduced-order control matrix \(\tilde{B}_k\).

Assuming we updated online SVD on snapshots \(\bar{X}_k\), we wish to inform reduced-order matrices \(\tilde{\bar{A}}_k\) and \(\tilde{P}_k\) about the change of rotation in scaled coordinate space (column space; the orthonormal basis of features). The change of rotation as new data becomes available can be tracked as \(\uis{K}{U'U}{k} = \tilde{U}_k^\top \tilde{U}_{k-1}\).

To align reduced-order matrices \(\tilde{\bar{A}}_k\) and \(\tilde{P}_k\) with this change in column space, first we decouple \(\uis{K}{U'U}{k}\) as follows:
\begin{equation}
    \uis{K}{U'U}{k} = \begin{bmatrix}
        \uis{K}{U'U}{k, p \times p} & \uis{K}{U'U}{k, p \times q} \\
        \uis{K}{U'U}{k, q \times p} & \uis{K}{U'U}{k, q \times q}
    \end{bmatrix},
\end{equation}

and then apply alignment to the reduced-order matrices as

\begin{align}
    \tilde{\bar{A}}_k & = \begin{bmatrix} \uis{K}{U'U}{k, p \times p} \tilde{A}_k \uis{K}{U'U}{k, p \times p}^\top & \uis{K}{U'U}{k, p \times p} \tilde{B}_k \uis{K}{U'U}{k, q \times q} \end{bmatrix} , \\
    \tilde{P}_k       & = {(\uis{K}{U'U}{k} P_k^{-1} \uis{K}{U'U}{k}^\top)}^{-1}.
\end{align}

What follows, is the update of reduced matrices \(\tilde{A}_k\) and \(\tilde{P}_k\) using truncated snapshots \(\tilde{X}_k\) and \(\tilde{X}^\prime_k\). The updates could be performed conveniently using proposed formulae in~\eqref{eq:precision-matrix-update} and~\eqref{eq:online-dmd-update} without modification.

While the computational cost of a mini-batch update is the same as applying the rank-one updates \(c\) times in the original formulation in Section~\ref{sec:online-dmd-updates}, the mini-batch updates of the proposed truncated online DMD yield significant benefits in computational cost.

\subsection{Hankel DMD}\label{sec:hankel-dmd}
Hankel DMD addresses several key problems in analyzing dynamical systems, particularly when dealing with certain complex, non-linear, or controlled systems with unknown time delays. The main idea is to construct a Hankel matrix from the data matrix \(X\) by embedding delay coordinates forming a Hankel matrix \(\bar{X}_{h, c}\). The Hankel matrix is then decomposed using DMD to find the low-rank representation of the system. Given snapshots \({\{x(t_i)\}}^c_{i=0}\), the \(h\)-times delayed embedding matrix \(\bar{X}_{h, c}\) of shape \(m + h \times c\) is formed as
\begin{equation}\label{eq:hankel}
    \bar{X}_{h, c} = \begin{bmatrix}
        x_0    & x_1       & \cdots & x_{c - h} \\
        \vdots & \vdots    & \ddots & \vdots    \\
        x_{h}  & x_{h + 1} & \cdots & x_{c},
    \end{bmatrix}
\end{equation}

which can be combined with rank-one updates by storing and vertically stacking snapshots \({\{x(t_i)\}}^{k+c}_{i=k}\) at each time step \(k\) and passing it to updates of DMD.~This will allow setting the larger number of time-delays, in case we wish to have \(h > c\). For particularly large systems with slow dynamics, we may specify delay steps \(h_d\) along with total time-delay to find a balance between computational cost and accuracy of capturing the system dynamics. This means that our embedding will be composed of snapshots \( [x_0, x_{h_d}, \ldots, x_h] \), sampled at the time intervals specified by the delay steps.

The updates of DMD, once again, employ~\eqref{eq:precision-matrix-update} and~\eqref{eq:online-dmd-update} providing time delayed embedding of snapshots pair \(\bar{X}_{h, c}\) and \(\bar{X}^\prime_{h, c}\).

\clearpage
\section{Method}\label{sec:method}
In this section, we introduce the change-point detection (CPD) algorithm based on subspace identification via online Dynamic Mode Decomposition (ODMD-CPD). The choice of subspace identification for CPD is motivated by the proven effectiveness of these methods in addressing complex problems (see Subsection~\ref{sec:subspace-based-cpd}). ODMD-CPD is applicable to non-linear, time-varying controlled systems with delays, where real-time data acquisition with irregular sampling is managed by a message queuing service. This approach is driven by real industrial challenges and grounded in the theoretical foundations discussed in Section~\ref{sec:preliminaries}. Here, we present our method coherently and provide a detailed description of the algorithm and guidelines for its application in subsequent sections.

\subsection{CPD-DMD}
As discussed in Section~\ref{sec:preliminaries}, the success of identifying a low-rank subspace over which the signal evolves, while removing noise terms, relies on selecting the appropriate rank for the subspace. Projecting data onto modes of this low-rank subspace can result in increased reconstruction error when non-conforming patterns appear in the data. Transient dynamics, in particular, cannot be adequately captured by the low-rank subspace~\citep{Kuehn2011, Gottwald2020}. Therefore, a valid selection of the subspace maximizes the reconstruction error for non-stationary signals and is crucial for its use in CPD~\citep{Moskvina2003}.

Long-term deployment in systems with time-varying characteristics connected to factors such as aging, wear, or environmental conditions necessitates sequential detection and updates to the subspace in a streaming manner. This allows the system to adapt to slow changes in the time series structure and to accommodate new operations that may persist for an undefined duration. The ODMD-CPD algorithm is designed to address these challenges, providing a robust and adaptive solution for CPD in time series data.

Firstly, when new snapshots are available, CPD-DMD updates the low-rank subspace over which the signal evolves. Secondly, the algorithm projects two stored windows of snapshot pairs, referred to as base and test matrices, onto the subspace to evaluate the reconstruction error. Finally, by comparing the reconstruction error between the base and test matrices, the algorithm computes the change-point statistics.

\subsection{Data Stream Management}
Efficient execution of the algorithm requires preprocessing incoming data streams and managing the history of snapshots to compute the change-point statistics. Algorithm~\ref{alg:preprocessing} shows a single pass of the data preprocessing and management procedure, described below. This procedure is executed for each available snapshot pair or in mini-batches of varying frequency and size. First, incoming snapshots are formed into time-delayed embeddings of a predefined number of delays \( h \), as shown in Eq.~\eqref{eq:hankel}. Next, the time-delayed embedding of one-step delayed snapshots \( X^\prime_{h, k: k + j} \) is compensated by control action if the control matrix \( B \) is known, or the time-delayed embedding \( X_{h, k: k + j} \) is augmented with control actions \( \Theta_{h, k: k + j} \) to form the augmented matrix \( \bar{X}_{h, k: k + j} \).

Four parameters \(a, b, c, d\) define three required snapshot sets; the base set \({\ui{\bar{X}}{B} = \{x_h(t_i)\}}^{k - b - c}_{i=k - a - b - c}\), the test set \(\ui{\bar{X}}{T} = {\{x_h(t_i)\}}^{k}_{i=k - c}\), and the learning pair \( \{\ui{\bar{X}}{L}, \ui{\bar{X}^\prime}{L}\} = {\{x_h(t_i), x_h(t_i^\prime )\}}^{k - b - c + j}_{i=k - b - c - d}\). Conveniently, storing snapshots pairs \( \{\ui{\bar{X}}{all}, \ui{\bar{X}^\prime}{all}\} = {\{x_h(t_i), x_h(t_i^\prime )\}}^{k}_{i=k - b - c - d}\) is sufficient to manage all required data efficiently. In Section~\ref{sec:guidelines}, we will explain the selection of these parameters.

\begin{algorithm}
	\caption{Single pass of data preprocessing and management procedure}\label{alg:preprocessing}
	\begin{algorithmic}[1]
		\REQUIRE{
			\(X_{k: k + j}\), 
			\(X^\prime_{k: k + j}\), 
			\(\Theta_{k: k + j}\), 
			\(\ui{\bar{X}}{all}\), 
			\(\ui{\bar{X}^\prime}{all}\), 
			\(h\), 
			\(B\), 
			\(a, b, c, d\) 
		}
		\ENSURE{
			\(\ui{\bar{X}}{L}\), 
			\(\ui{\bar{X}^\prime}{L}\), 
			\(\ui{\bar{X}}{B}\), 
			\(\ui{\bar{X}}{T}\), 
			\(\ui{\bar{X}}{all}\), 
			\(\ui{\bar{X}^\prime}{all}\), 
			\(j\) 
		}

		\STATE{\(j \gets \text{number of snapshots in } X_{k: k + j}\)}

		\STATE{\(X_{h, k: k + j} \gets \text{hankelize}(X_{k: k + j}, h)\)}
		\STATE{\(X^\prime_{h, k: k + j} \gets \text{hankelize}(X^\prime_{k: k + j}, h)\)}
		\STATE{\(\Theta_{h, k: k + j} \gets \text{hankelize}(\Theta_{k: k + j}, h)\)}

		\IF{\(B\) is known}
		\STATE{\(\bar{X}_{h, k: k + j} \gets X_{h, k: k + j}\)}
		\STATE{\(\bar{X}^\prime_{h, k: k + j} \gets X^\prime_{h, k: k + j} - B \Theta_{h, k: k + j}\)}
		\COMMENT{Eq.~\eqref{eq:control-compensation}}
		\ELSE{}
		\STATE{\(\bar{X}_{h, k: k + j} \gets \begin{bmatrix} X_{h, k: k + j} \\ \Theta_{h, k: k + j} \end{bmatrix}\)}
		\COMMENT{Eq.~\eqref{eq:augmented-matrix}}
		\STATE{\(\bar{X}^\prime_{h, k: k + j} \gets X^\prime_{h, k: k + j}\)}
		\ENDIF{}

		\STATE{\(\ui{\bar{X}}{L} \gets \ui{\bar{X}}{all}(k - b - c - d: k - b - c + j)\)}
		\STATE{\(\ui{\bar{X}^\prime}{L} \gets \ui{\bar{X}^\prime}{all}(k - b - c - d: k - b - c + j)\)}

		\STATE{\(\ui{\bar{X}}{all} \gets \begin{bmatrix} \ui{\bar{X}}{all}(j: k) & \bar{X}_{h, k: k + j} \end{bmatrix}\)}
		\STATE{\(\ui{\bar{X}^\prime}{all} \gets \begin{bmatrix} \ui{\bar{X}^\prime}{all}(j: k) & \bar{X}^\prime_{h, k: k + j} \end{bmatrix}\)}

		\STATE{\(\ui{\bar{X}}{B} \gets \ui{\bar{X}}{all}(k - a - b - c: k - b - c)\)}
		\STATE{\(\ui{\bar{X}}{T} \gets \ui{\bar{X}}{all}(k - c: k)\)}
	\end{algorithmic}
\end{algorithm}

\subsection{Learning Procedure}\label{learn-cpd}
The learning procedure involves updating the Dynamic Mode Decomposition (DMD) model with new snapshots and forgetting old ones to track the system's time-varying characteristics. This procedure is outlined in Algorithm~\ref{alg:learning}.

\begin{algorithm}
	\caption{Single pass of learning procedure of CPD-DMD}\label{alg:learning}
	\begin{algorithmic}[1]
		\REQUIRE{
			\(\ui{\bar{X}}{L}\),
			\(\ui{\bar{X}^\prime}{L}\),
			\(c\),
			\(j\)
		}
		\STATE{\(\tilde{c} \gets \text{number of snapshots in } \ui{\bar{X}}{L} \)}
		\STATE{\(j^\prime \gets \tilde{c} + j - c \)}
		\COMMENT{\(\tilde{c} \leq c\) smaller before fully loaded}
		\IF{\(j^\prime > 0\)}
		\STATE{\textbf{Revert:} DMD\( (
			-\ui{\bar{X}}{L}(: j^\prime),
			-\ui{\bar{X}^\prime}{L}(: j^\prime)
			) \)}
		\COMMENT{Eq.~\eqref{eq:precision-matrix-update},~\eqref{eq:online-dmd-update}}
		\ENDIF{}
		\STATE{\textbf{Update:} DMD\( (
			\ui{\bar{X}}{L}(- j: ),
			\ui{\bar{X}^\prime}{L}(- j: )
			) \)}
		\COMMENT{Eq.~\eqref{eq:precision-matrix-update},~\eqref{eq:online-dmd-update}}
	\end{algorithmic}
\end{algorithm}

Initially, we verify the number of snapshots in the learning set and revert the DMD subspace if the learning set is fully loaded. Note that the learning set might not be full at the start of the learning procedure but must contain at least \((m + l) h\) snapshots, assuming unique measurements and that the learning set \(\ui{\bar{X}}{L}\) has full column rank. Subsequently, we update the DMD subspace with new snapshots entering the learning set. This procedure is repeated for each snapshot pair available or in mini-batches, whose frequency and size may not be uniform, governed by the upstream message queuing service.

\subsection{Detection Procedure}\label{detect-cpd}
The detection procedure, executed before the learning procedure, computes the change-point statistics. This sequence is crucial to avoid false negatives, as new snapshots may represent transient dynamics, and updating the DMD subspace beforehand could result in misidentification. Although the impact of this sequence is minimal for rank-one updates since both procedures are executed in the same pass, its importance grows with the relative size of the mini-batch to the potential span of the change-point. Nevertheless, aligning with best practices prevents any information leaks.

Algorithm~\ref{alg:detection} details the detection procedure. First, we project the base and test matrices onto the DMD subspace. Second, we reconstruct the full state representation and calculate the sum of squared Euclidean distances between the data and their DMD reconstruction. Third, we normalize this sum by the number of snapshots in the matrices. Finally, we compute the change-point statistics as the ratio of errors between the test and base matrices.

In cases where the error ratio is less than 1, the reconstructed test set \(\ui{\tilde{\bar{X}}}{T}\) captures more information about \(\ui{\bar{X}}{T}\) than the reconstructed base set \(\ui{\tilde{\bar{X}}}{B}\) about \(\ui{\bar{X}}{B}\). This rare scenario typically occurs when the signal is stationary, but the noise variance decreases in the test set compared to the training set. Although this phenomenon is interesting as it indicates a change in noise variance and the end of transient regime states, it is not considered in this paper, and we truncate the value to 1 and shift the score to zero, defining the minimum energy of matching errors.

\begin{algorithm}
	\caption{Single pass of detection procedure of CPD-DMD}\label{alg:detection}
	\begin{algorithmic}[1]
		\REQUIRE{
			\(\ui{\bar{X}}{B}\),  
			\(\ui{\bar{X}}{T}\),  
			\( \Phi \)            
		}
		\ENSURE{
			\(Q_k\)  
		}
		\newline{\textbf{Project matrices onto DMD subspace}}
		\STATE{\(
			\ui{\tilde{\bar{X}}}{B} \gets \Phi^T \ui{\bar{X}}{B} \newline
			\ui{\tilde{\bar{X}}}{T} \gets \Phi^T \ui{\bar{X}}{T}
			\)}
		\newline{\textbf{Compute reconstruction errors}}
		\STATE{\(
			\ui{E}{B} \gets \sum_{i = k - a - b - c}^{k - b - c} \left \| \uis{\bar{X}}{B}{i} - \Phi \uis{\tilde{\bar{X}}}{B}{i} \right \|_F^2 \newline
			\ui{E}{T} \gets \sum_{i = k - c}^{k} \left \| \uis{\bar{X}}{T}{i} - \Phi \uis{\tilde{\bar{X}}}{T}{i} \right \|_F^2
			\)}
		\newline{\textbf{Normalize errors by the number of snapshots}}
		\STATE{\(
			\ui{E}{B} \gets \frac{\ui{E}{B}}{a} \newline
			\ui{E}{T} \gets \frac{\ui{E}{T}}{c}
			\)}
		\newline{\textbf{Compute change-point statistics}}
		\STATE{
			\(Q_k \gets \text{max}(0, \frac{\ui{E}{T}}{\ui{E}{B}} - 1)\)
		}
	\end{algorithmic}
\end{algorithm}

\subsection{Full Algorithm}
The complete CPD algorithm comprises three fundamental steps, as outlined in Algorithm~\ref{alg:cpd-dmd}. While the internal parameters of each step are abstracted for readability, their updates remain important. The proposed architecture is tailored for real-time execution, making it ideal for deployment in industrial environments characterized by dynamic data acquisition and irregular sampling patterns, often orchestrated by message queuing services.

\begin{algorithm}
	\caption{Single pass of CPD-DMD procedure}\label{alg:cpd-dmd}
	\begin{algorithmic}[1]
		\REQUIRE{
			\(X_{k: k + j}\),  
			\(X^\prime_{k: k + j}\),  
			\(\Theta_{k: k + j}\)  
		}
		\ENSURE{
			\(Q_k\)  
		}
		\newline{\textbf{Step 1: Preprocessing the data}}
		\STATE{\(
			\ui{\bar{X}}{L}, \ui{\bar{X}^\prime}{L}, \ui{\bar{X}}{B}, \ui{\bar{X}}{T} \gets
			\text{preprocessing}( X_{k: k + j}, X^\prime_{k: k + j}, \Theta_{k: k + j})
			\)}
		\COMMENT{Refer to Algorithm~\ref{alg:preprocessing}}
		\newline{\textbf{Step 2: Detecting change-points}}
		\STATE{\(
			Q_k \gets \text{detection}( \ui{\bar{X}}{B}, \ui{\bar{X}}{T} )
			\)}
		\COMMENT{Refer to Algorithm~\ref{alg:detection}}
		\newline{\textbf{Step 3: Updating the learning model}}
		\STATE{\(
			\text{learning}( \ui{\bar{X}}{L}, \ui{\bar{X}^\prime}{L} )
			\)}
		\COMMENT{Refer to Algorithm~\ref{alg:learning}}

	\end{algorithmic}
\end{algorithm}

\subsection{Guidelines}\label{sec:guidelines}
\begin{figure}[H]
	\centering
	\includegraphics[width=\linewidth]{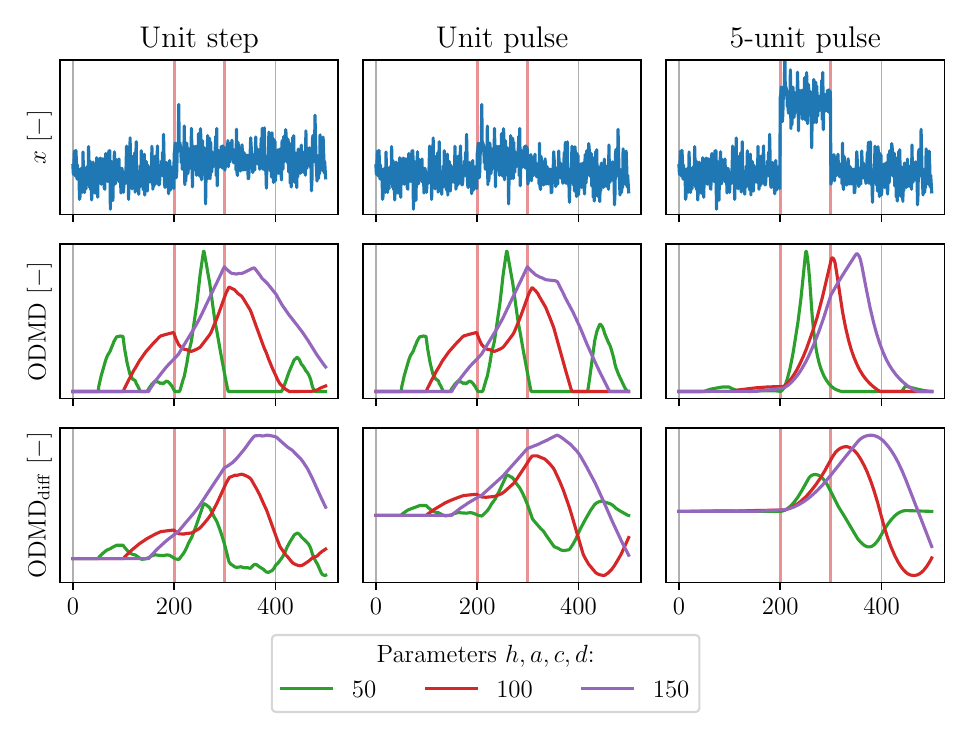}
	\caption{Increasing value of all the hyperparameters at once increases robustness to noise and delays peak of CPD statistics.}\label{fig:hyperparameters-all}
\end{figure}

\begin{figure}[H]
	\centering
	\includegraphics[width=\linewidth]{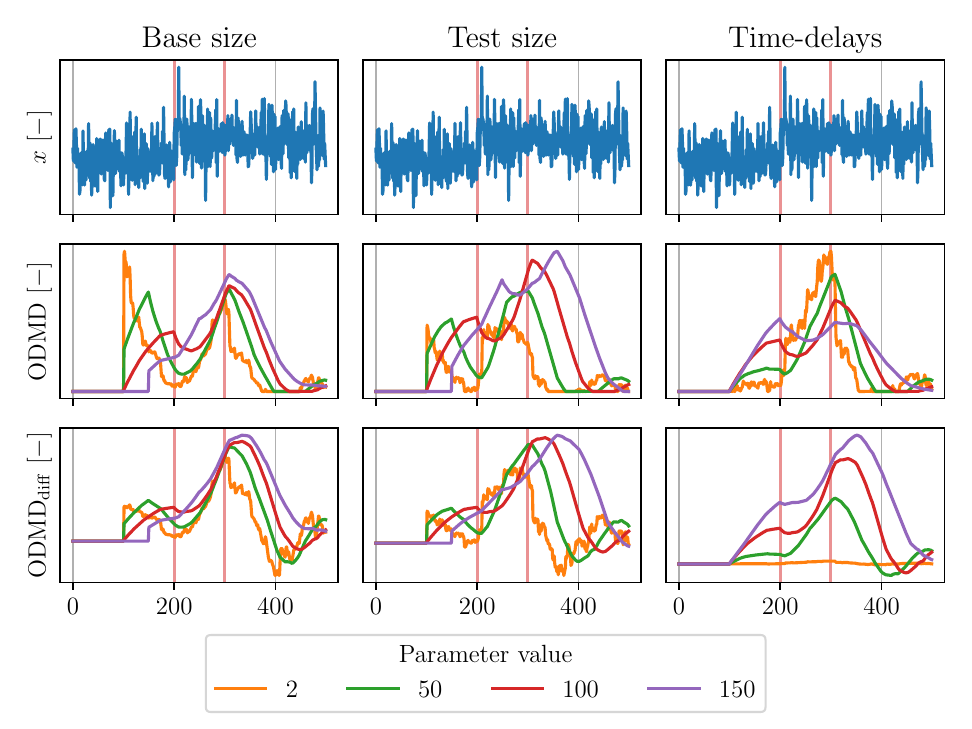}
	\caption{The influence of changing hyperparameter (denoted in the title of each column) values on CPD in synthetic unit step dataset.~Increasing value of base size stabilizes the score without significantly delaying the peak of CPD.~Increasing test size and number of time-delays in embedding increases robustness to noise more prominently while delaying the peak of CPD.}\label{fig:hyperparameters}
\end{figure}

This subsection aims to provide comprehensive guidelines for selecting hyperparameters tailored to specific use cases and problem types. Such guidance is essential for ensuring the tool's versatility across a broad spectrum of industrial applications characterized by unique conditions and specifications. By offering insights into the selection of hyperparameters, our guidelines aid the customization of the method to meet the specific requirements of different applications. To further demonstrate the impact of hyperparameter selection, we have included visual guidelines in Figure~\ref{fig:hyperparameters-all}, where all windowing parameters are changed at once, and Figure~\ref{fig:hyperparameters}, where the influence of base size, test size, and time-delays selection is shown.

\subsubsection{Approximating rank}
Determining the approximate rank \( r \) of the low-rank representation of the system is a crucial and inherently subjective step in any dimensionality reduction technique. To address this challenge, we recommend employing the systematic hard-thresholding algorithm proposed by Gavish and Donoho (2014) for extracting \( r \) from noisy data. This algorithm requires information about the ratio of the number of states \((m + l) h\) in the learning set \(\ui{\bar{X}}{L}\) and the learning window size \(a\), the selection of which is discussed later in Subsection~\ref{sec:base-window}.

Nevertheless, the proposed \( r \) may become computationally intractable for time-delayed embeddings. In such cases, we suggest using the row rank of the original data matrix \( m \) or, for augmented matrices, \( m + l \), in combination with hard-thresholding techniques to determine the optimal rank, particularly when linear system assumptions hold. For non-linear systems or situations where the collected data inadequately represent the underlying dynamics, a higher rank may be warranted to capture the dynamics effectively, while considering computational constraints and delayed response delivery. The online nature of the algorithm enables real-time adjustments to the rank, up to a certain threshold based on the significance of the \(r + 1\) singular value. These updates are facilitated by online singular value decomposition (SVD) algorithms, such as those presented in \citet{Brand2006} and \citet{Zhang2022}. For details on the implementation of rank-increasing updates, we refer readers to the original papers. Computational analysis by \citet{Zhang2019}, which applies for our proposed truncated version replacing number of states \(m\) with rank \(r\), indicates that the computational time of DMD updates scales with \( \mathcal{O}(r^2) \), with a number of floating-point multiplies of \(4n^2\), and memory requirements scale with \( \mathcal{O}(a r + 2 r ^ 2) \). This analysis can be utilized to evaluate the maximum rank for a given problem.

\subsubsection{Learning window size}
The learning window size \(d\) significantly influences the validity of the identified subspace and the accuracy of change-point detection. For identifying time-invariant systems, the learning window size should be sufficient to distinguish signal from noise and obtain the best approximation of the eigenvalues of the generating mechanism. For time-varying systems, the window size should encompass snapshots of single operating regimes or closely related operating regimes for effective learning. We propose setting the size of the base window as the lower bound on the learning window, although theoretically, the learning window could be smaller. The upper bound \(k >= b + c + d\) is determined by the number of available data points before the test window size \(c\) delayed by \(b\), as well as the size of the test window and the delay between the test and base windows. In summary, the learning window should be large enough to capture the system's dynamics without overlapping multiple operating states with distinct characteristics.

\subsubsection{Base window size and location}\label{sec:base-window}
The base window size \(a\) should reflect the expected duration of a stationary signal (single operation regime) within snapshots. This enables the reconstruction error of the base set to serve as a reference for the overall quality of the identified subspace and mitigates adverse effects on prediction accuracy. The base window should be located directly after the test window (\(b = 0\)). A value of \(a \leq d\) could prevent negative scores in collective anomalies, enabling effective differentiation between change points and collective changes.

\subsubsection{Test window size}
The test window size \(c\) determines the smoothness of change-point statistics over time. A larger test window results in smoother change-point statistics~\citep{Moskvina2003}. The selection of the test window size should consider the expected duration of the change-point, the nature of structural changes, and the desired smoothness of change-point statistics. Ideally, the peak of the statistics aligns precisely with the test window size delay from the change point. Smaller values of \(c\) decrease the delay, enhancing rapid detection but potentially missing slow drifts due to trends. Conversely, larger values of \(c\) increase stability and reduce false positives while potentially increasing false negative rates.

\subsubsection{Number of time-delays}
The number of time delays (\(h\)) is a critical parameter for ODMD-CPD in applications to non-linear systems with delayed effect of control action. Its selection relies on assumptions regarding the representativeness of snapshots with respect to the generating mechanism and the maximum expected delay of the control effect on system states. In the absence of such knowledge and with a reasonably large \(a\), \citet{Moskvina2003} recommend setting \(h = a / 2\) for the rank of the Hankel matrix. For larger \(a\), delay steps \(h_d\) may be used to capture the broadband dynamics of the system more effectively. The choice of \(h_d\) should be based on the maximum allowed number of features \((m + l)\frac{h}{h_d}\).

\subsubsection{Change-detection statistics threshold}
The threshold on change-detection statistics directly impacts the number of false positive or false negative alarms. Its proper selection further determines the delay of the detection alarm, as the rising change-point detection statistics cross the lower threshold sooner. As the CPD statistic, defined in Subsection~\ref{detect-cpd}, has no proper normalization and is influenced by the selection of other hyperparameters, the specification of constant values is challenging. As a general rule of thumb, higher recall (lower false negative rate) is achieved by setting the threshold lower. In contrast, higher precision (lower false positive rate) is achieved by setting the threshold higher. If accurate system tracking is achieved, i.e., \( r \) and \(h\) were selected so that the signal is extracted from noise well, the threshold could be set to zero while not compromising precision significantly. This may be desired in safety critical systems where higher recall helps to protect assets and life. Generally, the threshold should be set based on the desired trade-off between precision and recall, the change point's expected duration, and the structural changes' nature.

\clearpage
\section{Results}\label{sec:results}
This section presents the results of the proposed method applied to various datasets. The initial part of this section compares the proposed method with a related method for change point detection based on subspace identification using online SVD.~Artificial step detection highlights the significant differences in identifying subspaces using the two decomposition methods. Two real-world datasets are used to compare the performance of our proposed method with that of the alternative subspace identification method.

Secondly, we address the challenging real-world dataset involving faulty HVAC operation detection in an industrial-scale battery energy storage system (BESS). The final subsection compares the proposed method with other methods used for change-point detection on benchmarking data of a simulated complex dynamical system with control (CATS) and a laboratory water circulation system (SKAB).

\subsection{Artificial steps detection}
Change-point detection can be simplified to finding the temporal dynamics of a system not captured in data nor acknowledged by the supervisory control system. A simple artificial dataset with steps and Gaussian noise can validate the proper functioning of the proposed method. This dataset, initially proposed in~\citet{Kawahara2007}, highlights our proposed variation of online DMD as superior to online singular value decomposition.

The dataset consists of 10,000 snapshots with nine steps of increasing size and distance from the original operation point after every 1,000 snapshots. Gaussian noise is added to challenge one of the weaknesses of subspace-identification-based methods.

Our proposed method is compared to the online SVD-based CPD presented in \citet{Kawahara2007} using the same hyperparameters, as listed in Table~\ref{table:kawahara_comp_hyperparameters}.

\begin{table}[ht]
	\caption{Hyperparameters used for comparison with online SVD based CPD.}\label{table:kawahara_comp_hyperparameters}
	\centering
	\begin{tabular}{l S[table-format=3.0]}
		\toprule
		\textbf{Hyperparameter} & \textbf{Value} \\
		\midrule
		\(r\)                   & 2              \\
		\(a\)                   & 100            \\
		\(b\)                   & 0              \\
		\(c\)                   & 100            \\
		\(d\)                   & 300            \\
		\(h\)                   & 80             \\
		\bottomrule
	\end{tabular}
\end{table}

\begin{figure}[H]
	\centering
	\includegraphics[width=\linewidth]{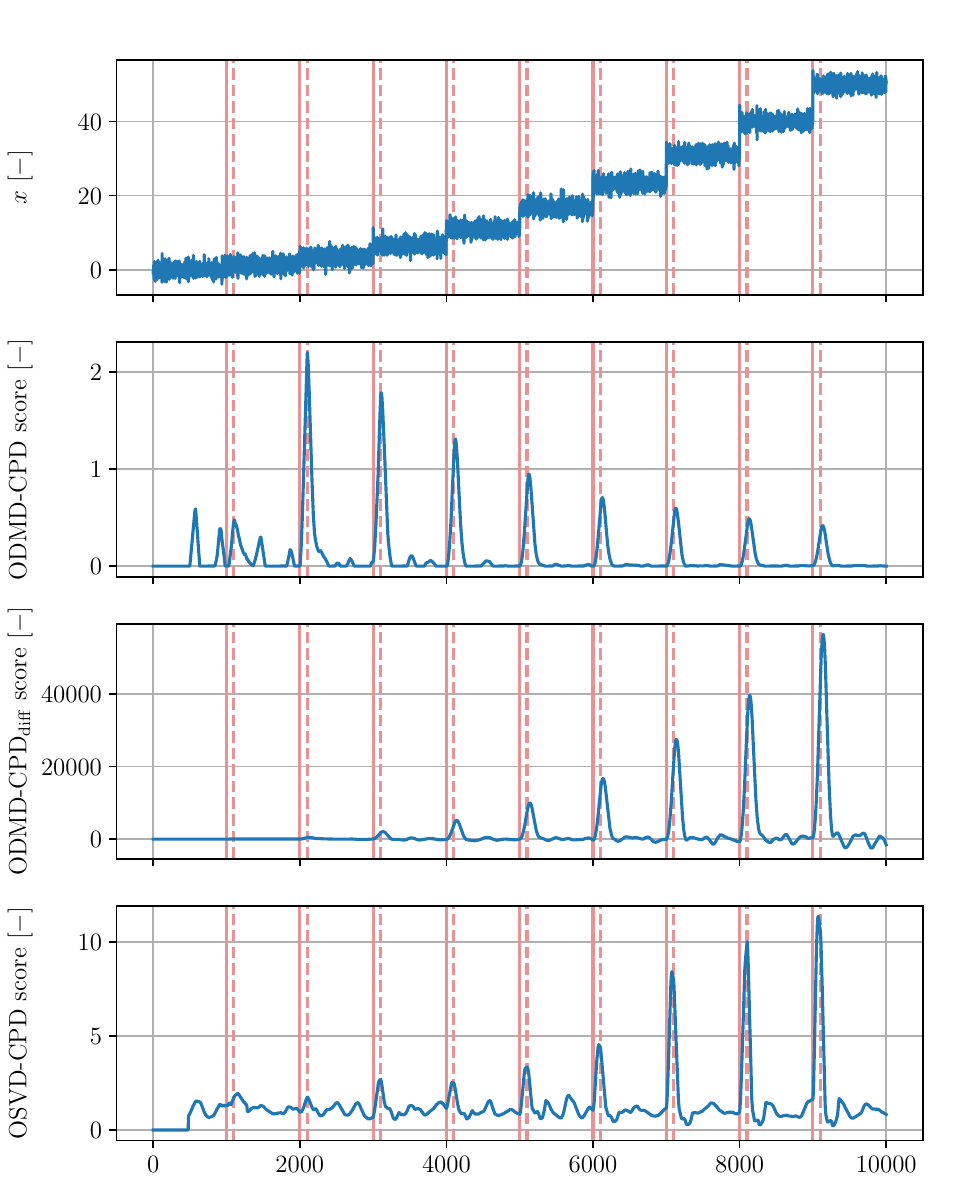}
	\caption{Steps detection in artificial data (1). Change score is evaluated for the proposed method as presented in Section~\ref{sec:method} (2), the proposed method evaluating score as the difference of errors (3), and the reference method using online SVD (4). Our method is capable of detecting minor CPs that are missed by the reference method.}\label{fig:artificial_steps_detection}
\end{figure}

The results in Figure~\ref{fig:artificial_steps_detection} show that while no method identified the first change-point at snapshot 1000, DMD-CPD discriminates minor change-points around the first operation point, with decreasing change-point statistics for subsequent detections. This decrease can be explained by the significant increase in absolute error between the original and reconstructed data for both the base and test sets, compared to their relative error, which decreases the residual of their division. Computing the difference between test and base set reconstruction errors gives evidence to the previous explanation. This relation of energy of the change-point detection to the actual dissimilarity of the data is a significant advantage of the proposed method over the deep learning techniques~\citep{DeRyck2021}. To sum up, the absolute dissimilarity of the data is reflected in the difference in the errors, while the relative dissimilarity is reflected in the error ratio.

The proposed method has a similar shape of statistics for error divergence to the error ratio statistics of the method proposed in \citet{Kawahara2007} but with significantly lower noise. In both cases, the peak of the change-point statistics is delayed by \(c\) snapshots, which is expected due to the nature of CPD-DMD.~The exact delay allows for precisely pinpointing the change-point time.

\subsection{Sleep Stage Detection via Respiration Signal}
Identifying change points in real-world data is challenging due to cyclic, seasonal, and environmental effects. In this context, we use a dataset of respiration signals from a sleep stage detection task. The datasets represent respiration (thorax extension), sampled at 10 Hz from different subjects. The data are manually labeled by Dr.~J. Rittweger from the Institute for Physiology at the Free University of Berlin. For details, refer to the original publication~\citep{Keogh2005}. For comparison with online SVD, we use the same hyperparameters as in the previous section, listed in Table~\ref{table:kawahara_comp_hyperparameters}.

\begin{figure}
	\centering
	\includegraphics[width=\linewidth]{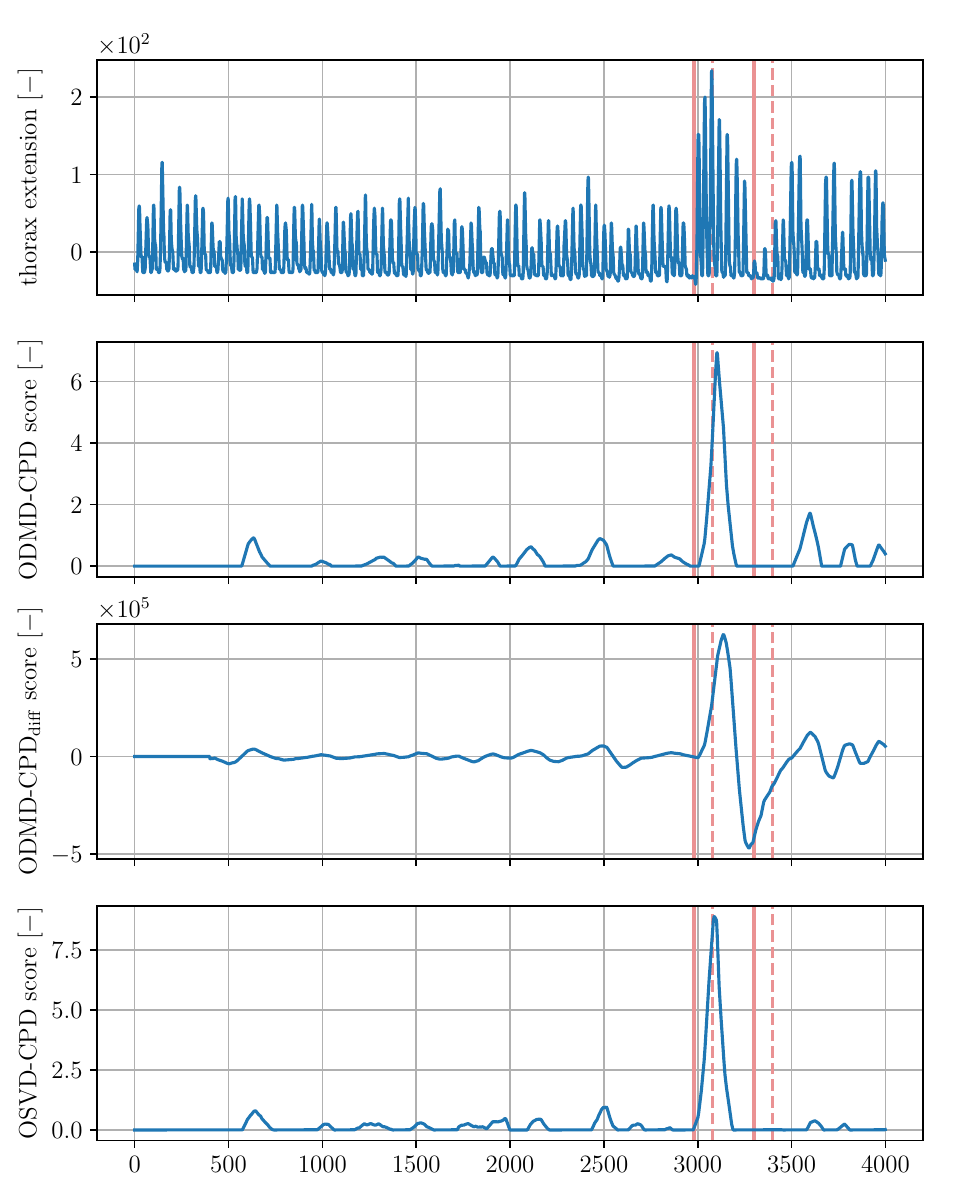}
	\caption{NPRS43: Sleep stage transition detection based on respiration data (1). Change score is evaluated for the proposed method as presented in Section~\ref{sec:method} (2), the proposed method evaluating score as the difference of errors (3), and the reference method using online SVD (4). While both methods detect the first CP, our method detects the second CP as well, albeit with a longer delay due to the proximity of the CPs.}\label{fig:nprs43}
\end{figure}

\paragraph{NPRS43}
The comparison results of experiments conducted on the NPRS43 dataset are presented in Figure~\ref{fig:nprs43}. This dataset spans approximately 7 minutes of sleep respiration signals of a subject. The subject is in stage II deep sleep for the first 5 minutes, then transitions through a short awake stage of approximately 1 minute to the stage I sleep. The transitions are marked as red vertical lines in the plot. Due to the size of the test set, the peak of change-point statistics is delayed by \(c\) snapshots and marked as red dashed vertical lines.

Both methods identify the first transition from stage II to the awake state with a peak of statistics delayed by \(c\). While SVD fails to recognize the second transition, our method displays an increased score with significant delay after the transition. Since the delay is longer than \(a + c\), it could be regarded as a false positive (FP) detection of a change point. Nonetheless, by visually examining data after the ground truth label, it could be argued that the second transition occurs more gradually and spans multiple breathing cycles, two with very short thrax extensions after the ground truth label, followed by two with larger thrax extensions and ended by two very large extensions. Our method seems to capture the middle point of this gradual transition as a reference, and learning windows cross the first transition point. The validity of this reasoning was not supported by the

\begin{figure}[H]
	\centering
	\includegraphics[width=\linewidth]{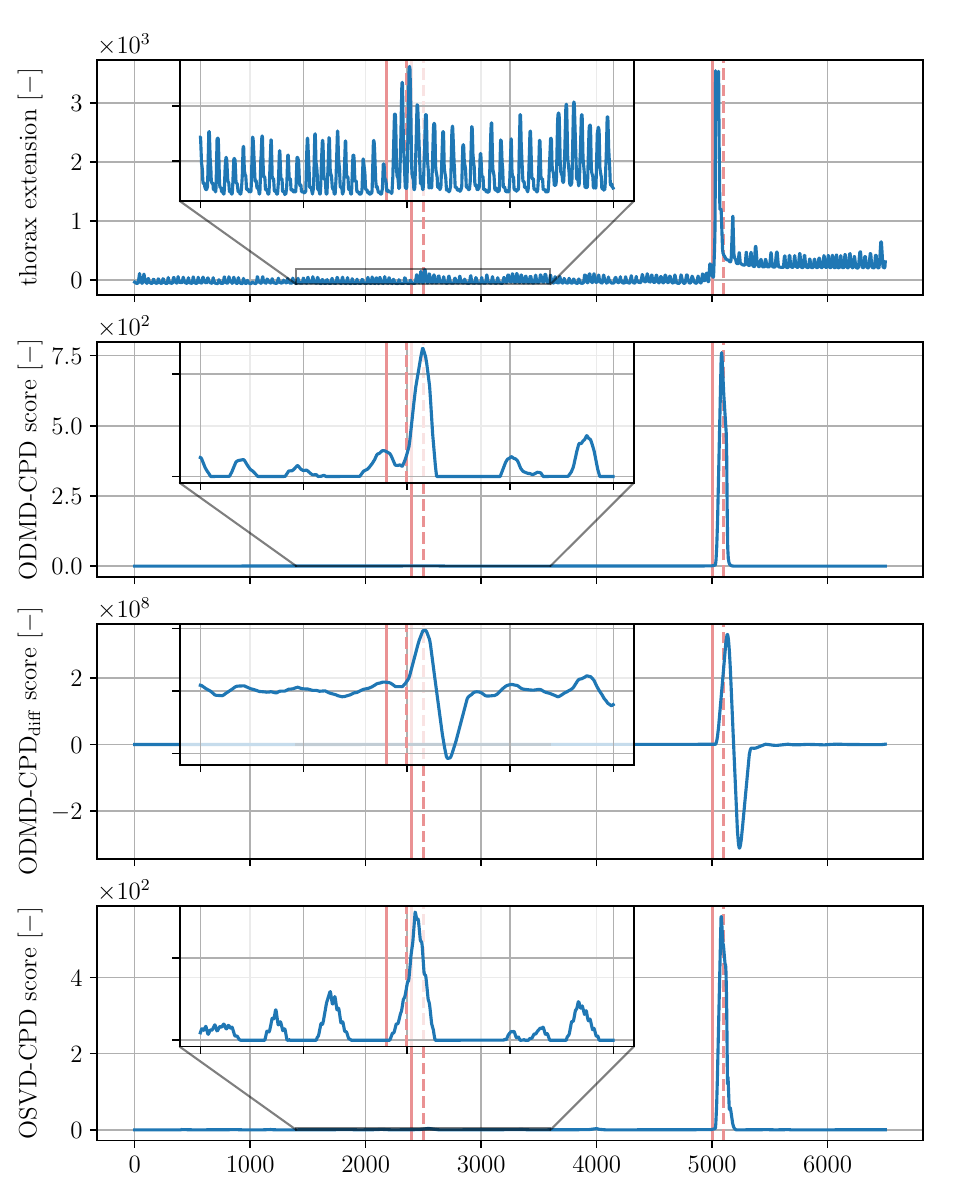}
	\caption{NPRS434: Sleep stage transition detection based on respiration data (1). Change score is evaluated for the proposed method as presented in Section~\ref{sec:method} (2), the proposed method evaluating score as the difference of errors (3), and the reference method using online SVD (4). While both methods detect CPs, our method detects the first one with a score three times higher than the peaks unrelated to tracked events, while OSVD only doubles the score.}\label{fig:nprs44}
\end{figure}

\paragraph{NPRS44}
The comparison results of experiments conducted on the NPRS44 dataset are presented in Figure~\ref{fig:nprs44}. This dataset spans approximately 11 minutes of sleep respiration signals of a subject. The subject is in stage II deep sleep for the first 4 minutes, then transitions through the stage I sleep, indicated by shallow breathing, for approximately 4 minutes to an awake state. The transitions are marked as red vertical lines, and the ideal change statistics peak as grey lines in the plot.

Both methods identify the transitions present in the dataset with high discrimination capacity. CPD-DMD has significantly fewer sharp peaks in areas where a transition is not anticipated compared to the SVD-based method. Moreover, CPD-DMD captures the second transition with higher prominence and achieves peak detection with a delay of exactly \(c\) snapshots. Under the same parametrization, CPD-DMD shows smoother scores and slightly better discrimination of the transitions.

\subsection{Simulated Two Tanks System with Input Delay}
To demonstrate the applicability of the proposed method to non-linear controlled systems with time delays, we demonstrate its performance on a simulated two-tank system with input delay represented by a system of ODE

\begin{equation}\label{eq:two_tanks_system}
	\begin{aligned}
		\diff{h_1(t)}{t} & = q(t-\tau) - \frac{k_1}{F_1}\sqrt{h_1(t)}                     \\
		\diff{h_2(t)}{t} & = \frac{k_1}{F_2}\sqrt{h_1(t)} - \frac{k_2}{F_2}\sqrt{h_2(t)},
	\end{aligned}
\end{equation}

where \(h_1(t)\) and \(h_2(t)\) are the levels in the tanks, \(q(t)\) is the control action, \(\tau \) is the time delay, \(k_1\) and \(k_2\) are the valve constants, and \(F_1\) and \(F_2\) are the cross-sectional areas of the tanks.

The system in Eq.~\ref{eq:two_tanks_system} is simulated with a sampling frequency of 0.1 Hz and 12,000 snapshots. After every 200 snapshots, a step change is introduced through the action of a valve selected randomly from the interval \(\interval[open left]{0.0}{0.03} m^3s^{-1}\). The time delay of the system's response to control is between 20 to 30 snapshots. The states of the system are subject to external stimuli and Gaussian noise with a variance of \(0.35\). After 4000 snapshots, the artificial sensor bias causes unit step change in the observation of tank levels for subsequent 1000 snapshots. Between 7600 and 8600, the system's response to control action is doubled, which may be caused by an offset in the control valve. Lastly, a linear trend is added to the system states between 9800 and 12000 snapshots to stimulate the increasing offset in the control.

The hyperparameters are selected based on the knowledge of the system dynamics. The learning window \(d\) is set to 2000 snapshots to capture the system's response to multiple control actions and different set points. The number of time delays in an embedding of states \(h\) is set to 200 snapshots to capture the system's dynamics, and \(\ui{h}{d}\) is set to 30 to increase performance while not significantly compromising representativeness of the dynamics. The time delays in a control action embedding are set to 30 to capture the control action responsible for the current system's response. As part of the on-the-flight preprocessing, we introduce a polynomial of degree 2 to the states to capture the non-linear dynamics of the system.~\(q\) and \(p\) are set to 2 and 1, respectively, to mitigate the impact of noise and capture information about the system's dynamics which time-delayed or polynomial embeddings might reveal.

\begin{figure}[H]
	\centering
	\includegraphics[width=\linewidth]{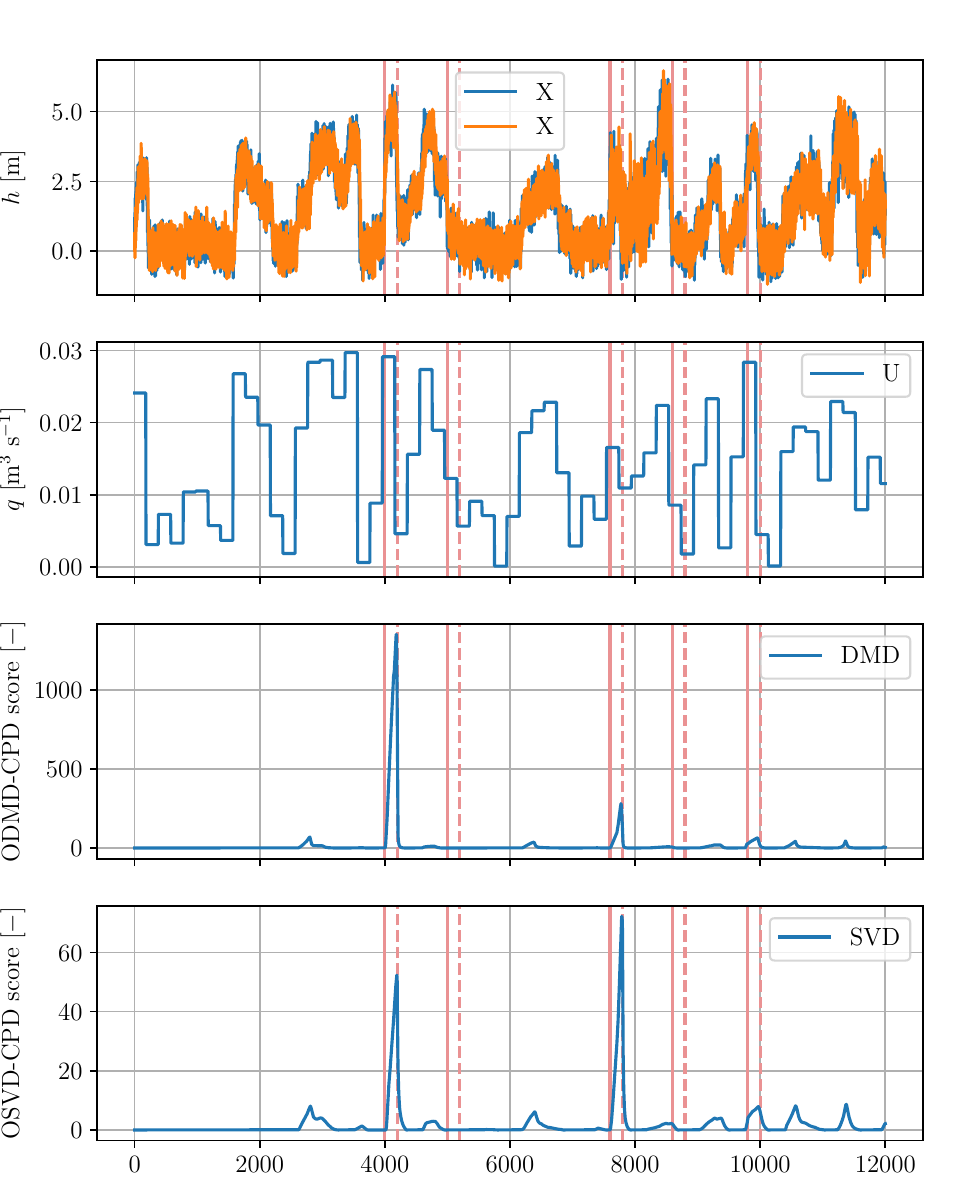}
	\caption{CPD detection on data of two tanks system (1) with delayed input (2) shows lower noise of DMD-CPD score (3) aiding recognition of the slow drift from snapshot 9800 as compared to the reference method (4).}\label{fig: non-linear}
\end{figure}

The results of the detection experiment on the simulated data, presented in Figure~\ref{fig: non-linear}, show that the proposed method accurately detects the starts of change points in the system's operation. The peak of change-point statistics is delayed by \(c\) snapshots, which is expected due to the nature of CPD-DMD. The proposed method detects all the change points in the system's operation, including the artificial sensor bias, the doubled response to control action, and the linear trend in the system states. While the SVD-based method detects all the change points as well, noisy CPD statistics may hinder the recognition of another change point regarding linear trends, which may be observed more often in real scenarios due to the aging of the device and seasonal effects.

\subsection{BESS --- Faulty HVAC Operation Detection}
This case study demonstrates detection performance on a real-world dataset of faulty HVAC operation in an industrial-scale battery energy storage system (BESS). The studied BESS has an installed capacity of 151 kWh distributed among ten modules with 20 Li-ion NMC cells. A hardware fault occurred on one of the module's cooling fans on 23rd August 2023 at 17:12:30. To protect the profitability of the BESS for the end user, the faulty BESS was operated securely until the fault was fixed. This case study aims to detect the transition from normal to faulty operation of the HVAC system based on temperature profile monitoring. The dataset is provided by the BESS operator and normalized to protect sensitive business information. It captures snapshots of six spatially distributed temperature sensors of the targeted BESS module operation at approximately 30-second intervals.

The selection of hyperparameters demonstrates the intuitiveness of parameter selection. The BESS is utilized in an industrial setting for availability time-shifting of energy generated by a solar power plant, subject to daily seasonality and weekly periodicity. The learning window \(d\) is set for 24 hours to reflect these patterns and track weekday and weekend operations well. The maximum C-rate of the BESS is 1.0, defining another important hourly time constant. With this knowledge, we can minimize the impact of charging events on change-point detection statistics; \(a\) and \(c\) window sizes are set to double the fastest charging rate, 240 samples, corresponding to 2 hours of operation. The number of time delays in the embedded matrix reflects the known dynamics of the system; hence, \(h\) is set to 240 samples.

\begin{figure}[H]
	\centering
	\includegraphics[width=\linewidth]{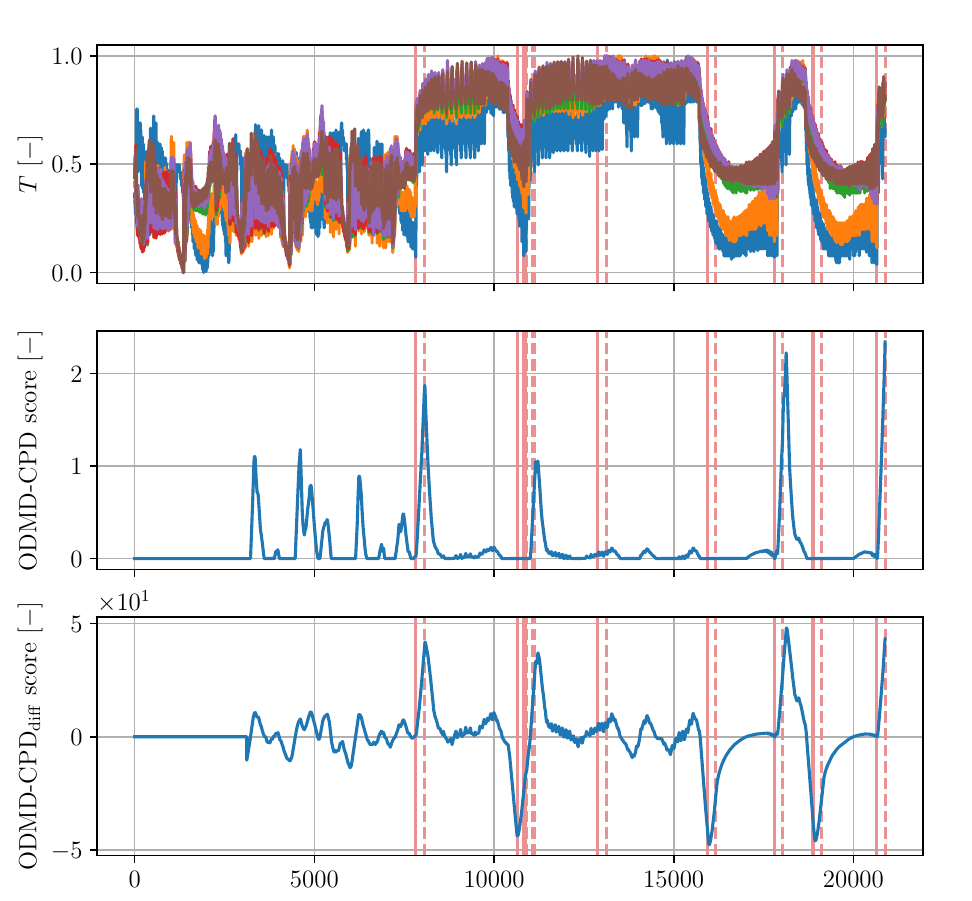}
	\caption{Faulty operation of HVAC in BESS results in altered operation temperature (1). Our proposed method detects a transition towards a novel state and displays multiple peaks in the CPD score (2). Meanwhile, an alternative formulation of our proposed method increases the prominence of the CPD score peaks during the transition towards the novel state and assigns negative peaks in the CPD score to the transitions back towards the original state (3).}\label{fig:bess}
\end{figure}

The results of the detection experiment on simulated data streaming from the BESS history replay, presented in Figure~\ref{fig:bess}, show that before the actual occurrence of the fault, the system detects multiple events of abnormal operation with a source other than the identified dynamical system from the data. The proposed method accurately detects the transition from normal to faulty operation of the HVAC system with high accuracy and low false positive rate, with the peak of change-point statistics delayed by \(c\) snapshots.

The proposed method detected three periods related to the transient normal behavior of the HVAC system, marked as change points. While it is challenging to confirm if the initial peaks in the CPD score were false positives, operators can interpret this information as a potential early warning of an impending fault. Such early warnings are valuable for detecting faults, preventing catastrophic consequences, and planning maintenance. The positions of the potential fault precursors match the detected anomalies in~\citet{Wadinger2024}, where detection was performed using an online anomaly detection method based on conditional probability distribution. This alignment supports cross-validation of both methods and lends credibility to identifying these events as fault precursors.

The evaluation of error difference shows that we can detect both the transition from normal to faulty operation and vice versa. Here, the peaks related to true anticipated CPs are more pronounced than those in the error ratio evaluation. This indicates the usefulness of the error difference evaluation in increasing the precision of the CPD.

\subsection{Laboratory Water Circulation System (SKAB)}
In this section, we compare the performance of the proposed method with commonly used change-point detection methods on a benchmark real-world dataset of a laboratory water circulation system (SKAB)~\citep{Katser2020}. The dataset represents a well-described industrial system with multiple sensors and well-defined operational and fault states characterized by collective anomalies or change points, as well as transitions between these states.

\citet{Katser2020} compare methods with default hyperparameters, which are listed in Table~\ref{table:comparison-models}, using the first 400 snapshots of each dataset as a training part. We follow the same procedure, and for CPD-DMD hyperparameters with task-specific tuning requirements, we use the training set of samples as a history of snapshots to establish the parameters.

\begin{table}[H]
	\caption{List of reference method and sources}\label{table:comparison-models}
	\centering
	\begin{tabular}{l l S[table-format=3.0]}
		\toprule
		\textbf{Algorithm} & \textbf{Source}       \\
		\midrule
		Conv-AE            & \citet{Pavithra2020}  \\
		Isolation forest   & \citet{Liu2008}       \\
		LSTM-AE            & \citet{Chollet2016}   \\
		MSCRED             & \citet{ZhangCh2019}   \\
		MSET               & \citet{Gross2000}     \\
		T-squared          & \citet{Hotelling1947} \\
		T-squared+Q (PCA)  & \citet{JoeQin2003}    \\
		Vanilla AE         & \citet{Chen2017}      \\
		Vanilla LSTM       & \citet{Filonov2016}   \\
		\bottomrule
	\end{tabular}
\end{table}

\begin{figure}[H]
	\centering
	\includegraphics[width=\linewidth]{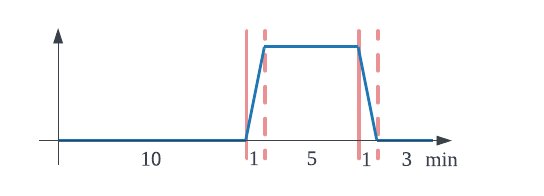}
	\caption{After 10 minutes of normal behavior, the system starts transitioning to a new operating point (indicated by the solid red line), which takes 1 minute to complete (indicated by the dashed red line). The system maintains this new operating state for 5 minutes, then transitions back to the original state over the next minute.}\label{fig:scab_interpretation}
\end{figure}

The evaluation is performed using NAB metrics presented in the work of~\citet{Ahmad2017}. These metrics operate over a window of snapshots. In the leaderboard proposed in~\citet{Katser2020}, the window is centered around the change point to establish metrics for reference methods. Nevertheless, from the definition of the change-point and the utilization of the window for scoring (please refer to the paper by~\citet{Lavin2015}), it is evident that the detector alerting a change-point half of the window size snapshots before the change-point actually occurs is considered perfect. Since the start of the transition towards the faulty state is marked as anomalous in the dataset, as seen in Figure~\ref{fig:scab_interpretation}, we stipulate that the detection before the start of the transition should be regarded as a false positive. Therefore, we modify the original evaluation metrics to observe the snapshots window after the change-point and evaluate the models' performance.

To ensure reproducibility and consistency, we created a fork of the original repository available at \url{https://github.com/MarekWadinger/SKAB}.

The results of experiments, presented in Table~\ref{table:skab_cpd_comparison-own}, show that the proposed method outperforms the reference methods in terms of standard NAB score, low FP, and low FN scores. Our proposed method and MSCRED have the lowest number of missed change points, making them more suitable methods for safety-critical systems where missed alarms may result in catastrophic consequences. Two variants of the proposed method are evaluated to show the influence of threshold \(t\) selection. Based on the results, we claim that the selection of threshold value, which is challenging for non-probabilistic approaches to CPD where CPD statistics have no proper normalization, may improve the FP score but does not significantly impact performance. Thus, this difficult-to-select parameter can initially be set to 0 and then increased to improve the false positive score when dealing with signals that are hard to reconstruct due to significant noise. Perhaps most interesting is the score of the perfect detector, which is not 100 as expected. This indicates that the standard NAB score does not guarantee a 100 for a perfect detector for various evaluation criteria, which is essential to consider when evaluating the relative performance of the methods with respect to the perfect detector under these criteria.

\begin{table}[H]
	\caption{Comparison of different algorithms based on NAB metrics. The best scores are highlighted.}\label{table:skab_cpd_comparison-own}
	\centering
	\begin{tabular}{l *3{S[table-format=3.2]}}
		\toprule
		\textbf{Algorithm}              &
		\multicolumn{1}{p{1.7cm}}{\centering \textbf{NAB}                                  \\ (standard)} &
		\multicolumn{1}{p{1.7cm}}{\centering \textbf{NAB}                                  \\ (low FP)} &
		\multicolumn{1}{p{1.7cm}}{\centering \textbf{NAB}                                  \\ (low FN)}
		\\
		\midrule
		Perfect detector                & 54.77          & 54.11          & 56.99          \\
		\midrule
		\textbf{CPD-DMD (\(t=0\))}      & \textbf{34.29} & 23.21          & \textbf{42.54} \\
		\textbf{CPD-DMD} (\(t=0.0025\)) & 33.43          & \textbf{23.28} & 41.71          \\
		MSCRED                          & 32.42          & 16.53          & 40.28          \\
		Isolation forest                & 26.16          & 19.50          & 30.82          \\
		T-squared+Q (PCA)               & 25.35          & 14.51          & 31.33          \\
		Conv-AE                         & 23.61          & 21.54          & 27.55          \\
		LSTM-AE                         & 23.51          & 20.11          & 25.91          \\
		T-squared                       & 19.54          & 10.20          & 24.31          \\
		MSET                            & 13.84          & 10.22          & 17.37          \\
		Vanilla AE                      & 11.41          & 6.53           & 13.91          \\
		Vanilla LSTM                    & 11.31          & -3.80          & 17.25          \\
		\midrule
		Null detector                   & 0.00           & 0.00           & 0.00           \\
		\bottomrule
	\end{tabular}
\end{table}

\subsection{Simulated Complex Dynamical System with Control (CATS)}
This section evaluates the performance of our method on the Controlled Anomalies Time Series (CATS) Dataset proposed in \citet{Fleith2023}. The dataset shows a simulation of a complex dynamical system with 200 injected anomalies, consisting of control commands, external stimuli, and 5 million snapshots of telemetry sampled at 1 Hz. While the generating mechanism is not described, the availability of the benchmark dataset, including the control action signals, makes it a good candidate for evaluating the proposed method. The dataset is meant for anomaly detection algorithms but contains numerous sequences of anomalous behavior. Compared to SKAB, this dataset has a far lower contamination level of 3.8\%, making it more suitable for evaluating the CPD performance, where events of change are underrepresented.

The evaluation uses the same metrics as the previous case study on a resampled dataset to 1-minute intervals, with a median taken for both features and targets. No public background on the generating mechanism complicated the hyperparameter selection. Based on the 58-day timespan captured in the dataset, we selected one day as the learning window and set the number of time delays to 4 hours, with a maximum limit of the final number of features set to 60. The reference window is double the size of the test window, 10 hours for the former and 5 hours for the latter. The ranks of the DMD were set to 10 and 4 for the states and control inputs, respectively.

The results of experiments, presented in Table~\ref{table:cats_cpd_comparison}, show that the proposed method outperforms all reference methods but MSCRED in terms of the standard NAB score evaluated on a 5-hour window starting to the right of the actual anomaly. While our method offers a significantly better FP score, reducing the number of false alarms, MSCRED offers a significantly better FN score. It is worth stating that while our proposed method and other reference methods completed the experiment within 1 hour, it took almost 24 hours for MSCRED. This means that it requires roughly one second per snapshot to process the data, which could challenge MSCRED's applicability in hard real-time scenarios with the original frequency of 1 Hz. In the given settings, our proposed method balances well between false positives and false negatives, with the lowest number of missed change points. The results indicate that the proposed method is capable of detecting change points in the complex dynamical system and can employ information about control inputs.

\begin{table}[H]
	\caption{Comparison of different algorithms based on NAB metrics. The best scores are highlighted.}\label{table:cats_cpd_comparison}
	\centering
	\begin{tabular}{l *3{S[table-format=3.2]}}
		\toprule
		\textbf{Algorithm}              &
		\multicolumn{1}{p{1.7cm}}{\centering \textbf{NAB}                                  \\ (standard)} &
		\multicolumn{1}{p{1.7cm}}{\centering \textbf{NAB}                                  \\ (low FP)} &
		\multicolumn{1}{p{1.7cm}}{\centering \textbf{NAB}                                  \\ (low FN)}
		\\
		\midrule
		Perfect detector                & 30.21          & 29.89          & 31.28          \\
		\midrule
		MSCRED                          & \textbf{37.19} & 13.46          & \textbf{47.18} \\
		\textbf{CPD-DMD (\(t=0\))}      & 25.66          & \textbf{20.62} & 29.84          \\
		\textbf{CPD-DMD}                & 17.84          & 15.01          & 20.06          \\
		Isolation forest (\(c=3.8\% \)) & 17.81          & 15.84          & 20.00          \\
		T-squared+Q (PCA)               & 11.80          & 11.40          & 12.30          \\
		LSTM-AE                         & 11.39          & 11.26          & 11.69          \\
		T-squared                       & 15.15          & 14.98          & 15.71          \\
		MSET                            & 14.48          & 13.43          & 15.60          \\
		Vanilla AE                      & 2.52           & 2.44           & 2.77           \\
		Vanilla LSTM                    & 0.73           & 0.70           & 0.82           \\
		Conv-AE                         & 0.15           & 0.14           & 0.18           \\
		\midrule
		Null detector                   & 0.00           & 0.00           & 0.00           \\
		\bottomrule
	\end{tabular}
\end{table}

\section{Conclusions}\label{sec:conclusions}
In this paper, we proposed truncation of online Dynamic Mode Decomposition with control and examined its efficacy in online subspace-based change point detection tasks. The approximation of subspace over which a complex system (possibly a non-linear time-varying controlled system with delays) evolves is traced using time-delayed embeddings created directly from the system's input-output non i.i.d.~streaming data. DMD enables the decomposition of the system's dynamics into a set of modes that can be used to reconstruct signals from the data, which are subject to noise and carry information about abrupt changes. The similarity of the original data to its reconstruction is evaluated over two windows: reference and test. The former establishes base reconstruction error, and the latter, which includes the latest snapshots provided by the streaming service, is evaluated for the presence of a change point. The size of the test window defines the delay of the peak CPD statistics, as shown on the synthetic dataset, and defines the maximum delay of the alarm under the assumption that the error crosses the selected threshold. The tradeoff between detection speed and the number of false positives can be tuned by changing this parameter. Although setting generally applicable default values of the proposed method's hyperparameters is impossible, we establish intuitive guidelines for their selection. We also show that while computing CPD statistics on error ratio reveals minor change points close to the origin, error divergence can be used to acquire statistics proportional to the actual difference. In the case study displaying real-world examples of faulty HVAC operation detection in BESS, we observe that the height of difference of the errors is proportional to the distance of the faulty state from normal operation. This is crucial for assessing the severity of deviations in the operation of industrial systems, which is relevant in overall risk assessment. In contrast, the error ratio hints at potential precursors of the transition towards the faulty operation. The proposed method is highly competitive, as shown on two benchmark datasets of a simulated complex system and a real laboratory system.

\section*{Notation}
\begin{longtable}{p{2cm}l}
	\toprule
	\textbf{Symbol}                & \textbf{Description}                                    \\
	\midrule
	\endhead{}
	\(^+\)                         & Moore-Penrose pseudoinverse                             \\
	\(^\top \)                     & matrix transpose                                        \\
	\(^{-1}\)                      & matrix inverse                                          \\
	\(|| . ||_F\)                  & Frobenius norm                                          \\
	\(\mathbf{0}_{r \times r}\)    & zeros matrix of shape \(r \times r\)                    \\
	\(A\)                          & state matrix                                            \\
	\(A_i\)                        & state matrix at \(i\)-th snapshot                       \\
	\(\tilde{A}\)                  & state matrix projected on POD                           \\
	\(\bar{A}_i\)                  & augmented state and control matrix at \(i\)-th snapshot \\
	\(B\)                          & input matrix                                            \\
	\(B_i\)                        & input matrix at \(i\)-th snapshot                       \\
	\(c\)                          & number of new snapshots                                 \\
	\(C\)                          & diagonal weights matrix of data matrix                  \\
	\(E\)                          & error between \(X\) and its projection                  \\
	\(E'\)                         & orthonormal basis of state projection error matrix      \\
	\(\Gamma \)                    & covariance matrix of posterior distribution             \\
	\(\textbf{I}_{r \times r}\)    & identity matrix of shape \(r \times r\)                 \\
	\(k\)                          & snapshot index                                          \\
	\(\ui{K}{E'E}\)                & orthogonal projection of error matrix                   \\
	\(l\)                          & number of control inputs                                \\
	\(\lambda \)                   & eigenvalue                                              \\
	\(\Lambda \)                   & diagonal matrix of eigenvalues                          \\
	\(m\)                          & number of states                                        \\
	\(p\)                          & rank of the reduced-order state matrix                  \\
	\(P_i\)                        & precision matrix at \(i\)-th snapshot                   \\
	\(\Phi \)                      & DMD modes                                               \\
	\(q\)                          & rank of the reduced-order control matrix                \\
	\(Q_i\)                        & lag covariance matrix of \(X_i\) and \(X'_i\)           \\
	\(r\)                          & number of modes                                         \\
	\(\tilde{\Sigma} \)            & singular values                                         \\
	\(t\)                          & alarm threshold on CPD statistics                       \\
	\(\theta \)                    & input at \(i\)-th snapshot                              \\
	\(\Theta \)                    & input matrix at \(i\)-th snapshot                       \\
	\(\tilde{U}\)                  & left singular vectors                                   \\
	\(\tilde{V}\)                  & right singular vectors                                  \\
	\( W \)                        & eigenvectors                                            \\
	\(x_i\)                        & state at \(i\)-th snapshot                              \\
	\(X\)                          & matrix of states                                        \\
	\(X_i\)                        & matrix of states at \(i\)-th snapshot                   \\
	\(X'_i\)                       & matrix of states at \((i+1)\)-th snapshot               \\
	\(\tilde{X}_i\)                & projected matrix of states at \(i\)-th snapshot         \\
	\(\bar{X}_i\)                  & matrix of states and controls at \(i\)-th snapshot      \\
	\(\ui{\tilde{\bar{X}}}{buff}\) & buffer of projected augmented matrices of states        \\
	\bottomrule
\end{longtable}

\section*{Additional information}
Our code and data are openly accessible on GitHub at the following URL:~\url{https://github.com/MarekWadinger/odmd-subid-cpd}.

\section*{CRediT authorship contribution statement}
\textbf{Marek Wadinger:} Conceptualization; Data curation; Formal analysis; Funding acquisition; Investigation; Methodology; Resources; Software; Validation; Visualization; Writing --- original draft; and Writing --- review \& editing.~\textbf{Michal Kvasnica:} Funding acquisition; Resources; Supervision; Validation~\textbf{Yoshinobu Kawahara:} Conceptualization; Project administration; Resources; Supervision; Validation.

\section*{Declaration of Competing Interest}
The authors declare that they have no known competing financial interests or personal relationships that could have appeared to influence the work reported in this paper.

\section*{Declaration of generative AI and AI-assisted technologies in the writing process}
During the preparation of this work the authors used Grammarly and GPT-3.5 in order to improve language and readability. After using this tool/service, the authors reviewed and edited the content as needed and take full responsibility for the content of the publication.

\section*{Acknowledgements}
The authors gratefully acknowledge the contribution of the Program to support young researchers under the project Adaptive and Robust Change Detection for Enhanced Reliability in Intelligent Systems. The authors gratefully acknowledge the contribution of the Scientific Grant Agency of the Slovak Republic under the grant 1/0490/23 and grant VEGA `1/0691/21'. This research is funded by the Horizon Europe under the grant no. 101079342 (Fostering Opportunities Towards Slovak Excellence in Advanced Control for Smart Industries).

\bibliographystyle{elsarticle-harv}
\bibliography{main}

\end{document}